\documentclass[lettersize,journal]{IEEEtran}
\usepackage{amsmath,amsfonts}
\usepackage{multirow}
\usepackage{algorithmic}
\usepackage{algorithm}
\usepackage{array}
\usepackage[caption=false,font=normalsize,labelfont=sf,textfont=sf]{subfig}
\usepackage{textcomp}
\usepackage{stfloats}
\usepackage{url}
\usepackage{verbatim}
\usepackage{graphicx}
\usepackage{cite}
\usepackage[colorlinks]{hyperref}
\begin{document}

\title{HyperLips: Hyper Control Lips with High Resolution Decoder for Talking Face Generation}

\author{
	Yaosen Chen, 
	Yu Yao, 
	Zhiqiang Li, 
	Wei Wang, 
	Yanru Zhang, 
	Han Yang, 
	Xuming Wen 
	
	\thanks{Yaosen Chen is with University of Electronic Science and Technology of China, Chengdu, Sichuan, 611731 China, and also with Media Intelligence Laboratory, ChengDu Sobey Digital Technology Co., Ltd, Chengdu, Sichuan, 610041 China  e-mail: (chenyaosen@sobey.com).}
	
	\thanks{Yu Yao and Zhiqiang are with  Media Intelligence Laboratory, ChengDu Sobey Digital Technology Co., Ltd, Chengdu, Sichuan, 610041 China, e-mail:(yaoyu, lizhiqiang@sobey.com).}
	
	\thanks{Wei Wang is with Media Intelligence Laboratory, Chengdu Sobey Digital Technology Co., Ltd, Chengdu, Sichuan, 610041, China, and also with Peng Cheng Laboratory, Shenzhen 518055,China e-mail:(wangwei@sobey.com).}
	
	\thanks{Yanru Zhang is with University of Electronic Science and Technology of China, Chengdu, Sichuan, 611731 China e-mail:(yanruzhang@uestc.edu.cn).}

	\thanks{Han Yang is with University of Electronic Science and Technology of China, Chengdu, Sichuan, 611731 China, and also with  Media Intelligence Laboratory, ChengDu Sobey Digital Technology Co., Ltd, Chengdu, Sichuan, 610041 China  e-mail: (yanghan@sobey.com).}
	
	\thanks{Xuming Wen is with Media Intelligence Laboratory, Chengdu Sobey Digital Technology Co., Ltd, Chengdu, Sichuan, 610041, China, and also with Peng Cheng Laboratory, Shenzhen 518055,China e-mail:(wenxuming@sobey.com).}
	
}

\markboth{Journal of \LaTeX\ Class Files,~Vol.~14, No.~8, August~2023}%
{Shell \MakeLowercase{\textit{et al.}}: A Sample Article Using IEEEtran.cls for IEEE Journals}


\maketitle

\begin{abstract}
Talking face generation has a wide range of potential applications in the field of virtual digital humans. However, rendering high-fidelity facial video while ensuring lip synchronization is still a challenge for existing audio-driven talking face generation approaches. To address this issue, we propose HyperLips, a two-stage framework consisting of a hypernetwork for controlling lips and a high-resolution decoder for rendering high-fidelity faces.
In the first stage, we construct a base face generation network that uses the hypernetwork to control the encoding latent code of the visual face information over audio. 
First, FaceEncoder is used to obtain latent code by extracting features from the visual face information taken from the video source containing the face frame.Then, HyperConv, which weighting parameters are updated by HyperNet with the audio features as input, will modify the latent code to synchronize the lip movement with the audio. Finally, FaceDecoder will decode the modified and synchronized latent code into visual face content. 
In the second stage, we obtain higher quality face videos through a high-resolution decoder. To further improve the quality of face generation, we trained a high-resolution decoder, HRDecoder, using face images and detected sketches generated from the first stage as input.
Extensive quantitative and qualitative experiments show that our method outperforms state-of-the-art work with more realistic, high-fidelity, and lip synchronization.  Project page: \href{https://semchan.github.io/HyperLips_Project/}{https://semchan.github.io/HyperLips\_Project/}
\end{abstract}


\section{Introduction}
\IEEEPARstart{W}{ith} the growth of audio-visual content~\cite{toshpulatov2023talking,chen2021boundary,chen2021capsule,chen2022video,xie2007realistic,yu2021multimodal} and the rise of the metaverse, talking face generation has broad application prospects in visual dubbing~\cite{kr2019towards, prajwal2020lip,xie2021towards}, digital assistant~\cite{thies2020neural}, virtual human~\cite{ravichandran2023synthesizing}, animation film and other fields, and has attracted more and more attention. 

\begin{figure}[t]
	\centering
	\includegraphics[width=0.9\linewidth]{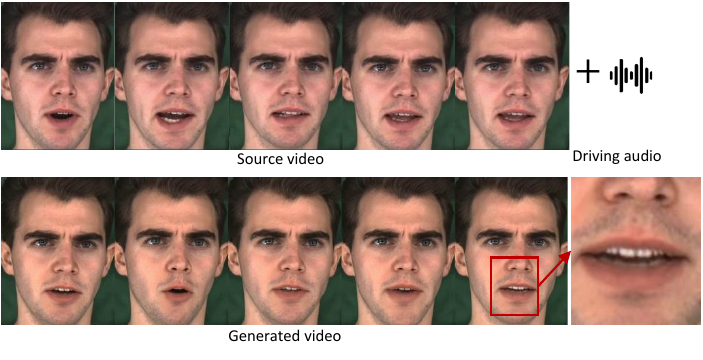}
	\vspace{-9pt}
	\caption{ Given the visual face information of source videos (upper left) and driving audio (upper right), our method is capable of rendering and generating more realistic, high-fidelity, and lip-synchronized videos (lower). See the zoom-in patches, our method can see details such as teeth.}
	\vspace{-15pt}
	\label{fig:intro}
\end{figure}

Based on the input requirements of the application, talking face generation methods can be categorized as driving audio only~\cite{tang2022real,guo2021ad,ye2023geneface,ye2023geneface++,chatziagapi2023lipnerf}, driving audio with a single frame~\cite{chen2019hierarchical,zhang2023sadtalker,eskimez2021speech}, and  driving audio with source video (or multiple frames)~\cite{prajwal2020lip,park2022synctalkface,zhong2023identity,zhang2023dinet,ye2022audio} types. 
For driving audio only, it is person-specific primarily and requires re-training for videos captured by the target speaker. For example, using a neural radiance field to train the implicit 3D representation of a captured video of a specific speaking person can observe the person's speech in a novel view~\cite{tang2022real,guo2021ad}, but the rendering results always look unnatural during movement. Due to the lack of facial and motion information as input for driving audio with a single frame, although some studies have done enough work, it is still impossible to generate accurate expressions and natural motion sequences~\cite{zhang2023sadtalker}.
For driving audio with source video, as shown in Fig.~\ref{fig:intro},  the expressions and movements of the characters in the generated video  are mostly taken from the source video, which naturally has realistic expressions and natural movements. 
In this case, there are two main challenges: 1. How to produce more accurate \textbf{lip synchronization} in generated videos; 2. How to render more \textbf{high-fidelity} faces, especially high-definition lips and teeth, in generated videos.
To produce more accurate {lip synchronization} in generated videos, Wav2lip~\cite{prajwal2020lip} proposed a lip sync discriminator to improve the performance of lip synchronization in unconstrained videos; SyncTalkFace~\cite{park2022synctalkface} proposed an audio lip memory that uses visual information of the mouth region corresponding to the input audio and enforces fine-grained audio-visual coherence; IP\_ LAP~\cite{zhong2023identity} leveraged a transformer-based landmark generator to infer lip and jaw landmarks from the audio to synchronize lip shape. These methods typically fuse the visual and audio features before decoding. However, the dimensions of audio and visual features are different, so additional processing is required to make them the same size for feature fusion.
In the first stage of our method, we encode visual face information as a latent code, then modify the latent code by a HyperConv convolution operation, and finally decode the modified latent code into visual face content. The weight parameters of HyperConv are generated by constructing a hypernetwork using audio features as input, thus achieving audio control of  lip movement in the rendered visual content. We use hypernetwork to avoid additional operations during the fusion of visual and audio features and to ensure lip synchronization in the generated videos better. 
Our idea is similar to the Audio Conditioned Diffusion Model~\cite{bigioi2023speech}, which takes audio information as the condition variable. Still, this method takes the diffusion model as the network architecture, which increases the demand for computational resources.


To render more high-fidelity faces, DINet~\cite{zhang2023dinet} proposed a Deformation Inpainting Network to achieve face visually dubbing on high-resolution videos, but it may generate artifacts out of face if mouth region covers background. IP\_LAP~\cite{zhong2023identity} leverage the prior appearance information which is extracted from the lower-half occluded target face and static reference images, it may fail when landmark cannot be detected in the reference images. Another possible method is to increase the input resolution based on networks such as Wav2lip~\cite{prajwal2020lip} or SyncTalkFace~\cite{park2022synctalkface}, but this not only increases the need for training resources but also does not render well, resulting in persistent artifacts.
In the second stage of our method, we propose a high-resolution decoder (HRDecoder) to further optimize the fidelity of generating faces. We trained the network using the facial data generated in the first stage and the corresponding facial sketches, guided by the sketches, to achieve facial enhancement. 

In summary, the contributions of our work are as follows:

\begin{itemize}
	\item We propose a hypernetwork based on audio information to control the generation of facial visual content that improves lip synchronization.
	\item We propose a high-resolution decoder with facial sketch guidance that can render more high-fidelity faces, especially high-definition lips and teeth, in generated videos.
	\item Extensive experiments show that our method can achieve significantly better talking face generation performance in terms of lip synchronization and face quality.
\end{itemize}

\section{Related Work}

\subsection{Audio-Driven Talking Face Generation}

In methods that only use audio input for audio-driven talking face generation~\cite{tang2022real,guo2021ad,ye2023geneface,ye2023geneface++,chatziagapi2023lipnerf}, collecting audio and video for person-specific and re-training is usually necessary. By introducing a neural radiance field (NeRF)~\cite{mildenhall2021nerf} to represent the scenes of talking heads~\cite{guo2021ad}, it can be controlled to render the face in a novel view. RAD-NeRF~\cite{tang2022real}  decompose the inherently high-dimensional talking portrait representation into three low-dimensional feature grids, that makes can rending the talking portrait in real-time. GeneFace~\cite{ye2023geneface} propose a variational motion generator to generate accurate and expressive facial landmark and uses a NeRF-based renderer to render high-fidelity frames. Due to a lack of prior information, these tasks still struggle to render realistic expressions and natural movements.
To drive a single facial image, ATVGnet~\cite{chen2019hierarchical} devises a cascade GAN approach to generate a talking face video, which is robust to different face shapes, view angles, facial characteristics, and noisy audio conditions. Recently, SadTalker~\cite{zhang2023sadtalker} propose a novel system for a stylized audio-driven single-image talking face animation using the generated realistic 3D motion coefficients, improving motion synchronization and video quality, but it is still impossible to generate accurate expressions and natural motion sequences.
The method of driving audio with source video is the most competitive because it can provide enough realistic facial expressions and natural movement information.  Wav2lip~\cite{prajwal2020lip}, SyncTalkFace~\cite{park2022synctalkface}, IP\_LAP~\cite{zhong2023identity}, DINet~\cite{zhang2023dinet} all belong to this category, mainly focusing on how to generate better lip synchronization and higher fidelity faces.
\subsection{HyperNetwork}
Hypernetwork~\cite{2017HyperNetworks} was originally proposed to generate the weights for a larger network. In evolutionary computation, operating directly on large search spaces consisting of millions of weight parameters is difficult. A more efficient method is to evolve a smaller network to generate the weight structure for a larger network, so that the search is constrained to the much smaller weight space. The idea of weight generation is easy to use for controllable generation tasks. Chiang et al. \cite{chiang2022stylizing} leverage it to control the style of the 3d scene representation. UPST-NeRF~\cite{chen2022upst} uses hypernetwork to contorl the universal photorealistic style transfer for 3D scene. 
With the rise of Large Language Models (LLMs)~\cite{radford2018improving,radford2019language,brown2020language,peng2023instruction} and generative models~\cite{rombach2022high}, hypernetwork has also become one of the necessary skills for fine-tuning LLMs. 
Essentially, the idea of generating weight parameters in a hypernetwork to control the large network method is similar to the Audio Conditioned Diffusion Model ~\cite{bigioi2023speech} and the Conditioning Mechanisms in Latent Diffusion Models~\cite{rombach2022high}, both of which achieve controllable output of the decoding through a control variable. In our method, however, we perform controllable generation relatively simply rather than using the diffusion model for generation.
Wang et al~\cite{wang2021regularization} use hypernetwrok for the application of magnetic resonance imaging reconstruction.
\begin{figure*}[t]
	\centering 
	\includegraphics[width=0.9\linewidth]{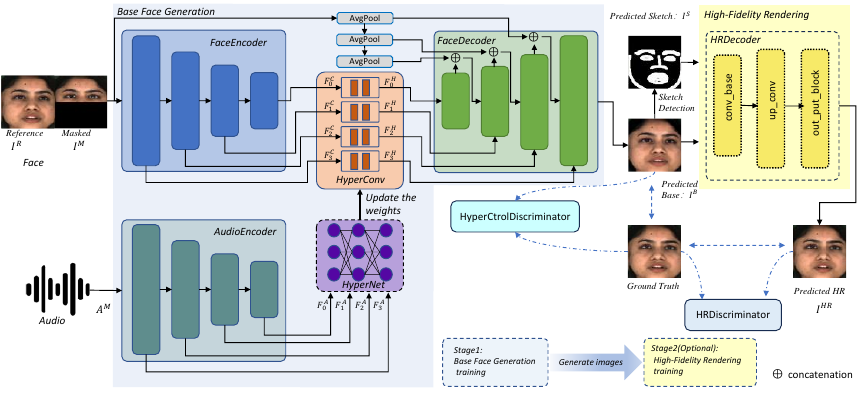}
	\caption{Overview of our proposed model.~It can be divided into two stages: 
		\textbf{(1) Base Face Generation}. FaceEncoder encode the visual face information (Reference and Masked) as a latent code, then modify the latent code by a HyperConv convolution operation, and finally FaceDecoder decode the modified latent code into visual face content. The weight parameters of HyperConv are updated by a hypernetwork using audio features as input, thus achieving audio control of  lip movement in the rendered visual face content (Base Face).
		\textbf{(2) High-Fidelity Rendering} . The high-resolution decoder (HRDecoder) is used to further optimize the fidelity of generating faces. We trained the network using the facial data generated in the first stage and the corresponding facial sketches, guided by the sketches, to achieve facial enhancement. Therefore, the input to HRDecoder is the concatenation feature of the base face with the sketch extracted from the base face, and the output is the high-fidelity face.
	}
	\label{fig:framework}
\end{figure*}
\subsection{Prior Based Face Restoration}
Face restoration is to restore the high-quality face image from the degraded face image~\cite{wang2022survey}. Face restoration is divided Non-prior and Prior based methods.
FSRNet~\cite{chen2018fsrnet} uses a coarse Super-Resolution (SR) network to recover coarse images, which are then processed by a fine SR encoder and a prior face-passing map estimation network, respectively.
Finally, image features and prior information are fed to the fine SR decoder to get the results.
In ~\cite{shen2018deep}, it uses the semantic label as the face prior. The semantic label is extracted from the input image or coarse deblurred image by a face parsing network. The final sharp image is generated from a deburring network with the input of the concatenation of the blurred image and the face semantic label.
Yin et al. ~\cite{yin2020joint} propose a joint alignment and face super-resolution network to learn landmark localization and face restoration jointly.
In our work, we use the landmark sketches detected from the relatively low-quality face generated in the first stage as input to guide the HRDecoder to achieve face enhancement to render high-fidelity faces.

\section{Proposed Method}
The overview of our framework is shown in Fig.~\ref{fig:framework}. We aim to generate a high-fidelity talking face video with synchronized lip movements by implementing the occluded face in the lower half of the input video frame by frame, given an audio and video sequence. Our proposed method consists of two stages: Base Face Generation and High Fidelity Rendering. In Base Face Generation, we designed a hypernetwork that takes audio features as input to control the encoding and decoding of visual information to obtain base face images. In high-fidelity rendering, we trained an HRDecoder network using face data from the network trained in the first stage and corresponding face sketches to enhance the base face.

\subsection{Base Face Generation}
\subsubsection{Hyper Control Lips}
Given the reference image ${I}^{R} \in \mathbb{R}^{3\times\ H^{I} \times W^{I}}$ and the masked image (occluded face in the lower half of the reference image) $I^{M} \in \mathbb{R}^{3\times\ H^{I} \times W^{I}}$, FaceEncoder obtains the latent code $L^{C} = \{F^{C}_{i}\}|0 \leq i \leq 3$ by extracting the $I^{R}$ and $I^{M}$ concatenation as inputs. Next, we use HyperConv's convolution operation to process $L^{C}$ and obtain $L^{H} = \{F_{i}^{H}\}|0 \leq i \leq 3$. Finally, we decoded $L^{H}$ with FaceDecoder to get the predicated base face $I^{B} \in \mathbb{R}^{3\times\ H \times W}$. The HyperConv implicit function can be formulated as follows:
\begin{equation}
	\begin{aligned}
		\mathcal{F}_{\Theta}:(L^{C})  \rightarrow (L^{H}),
	\end{aligned}
\end{equation}
where $\Theta$ is the weight parameter of the HyperConv convolution operation, which predicated by the HyperNet. HyperNet is composed of MLP, with audio deep features  $\{F_{i}^{A}\}|0 \leq i \leq 3$ extracted by AudioEncoder as input. 
For the input of AudioEncoder, we follow ~\cite{prajwal2020lip} to extract the Mel-spectrogram of the audio as ${A}^{M} \in \mathbb{R}^{\ H^{A} \times W^{A}}$.

The size of the audio mel-spectrogram is usually $16 \times 80$, i.e., $H^{A}$=16 and $W^{A}$=80. In contrast, the image size is usually not the same as the size of the audio mel-spectrogram, i.e., $H^{I}$=$W^{I}$=128 as default in our method.
In our method, we do not need to unify the dimensions of audio features and visual features, so no additional operations are required compared to other methods.

\subsubsection{Loss Function for Base Face Generation}
To be competitive in lip synchronization and fidelity, we constrain the generated base face by integrating multiple losses. We adopt its architecture from~\cite{prajwal2020lip} to design a quality discriminator which we called HyperCtrolDiscriminator, indicates as $\mathcal{D}^{B}$. 
Thereform, the overall generation process can be formulated as follows:
\begin{equation}
	\begin{aligned}
		I^{B}=\mathcal{G}^{B}((I^{R}\oplus I^{M}),A^{M}),
	\end{aligned}
	\label{eq:gb}
\end{equation}
where, $\oplus$ indicate concatenation.
In this way, we can consider the base face generation as a generator,indicates as $\mathcal{G}^{B}$, consisting of the following modules: FaceEncoder, HyperConv, FaceDecoder, AudioEncoder and HyperNet. We train the discriminator by adding the following loss:
\begin{equation}
	\begin{aligned}
		\mathcal{L}^{B}_{d}={\mathbb{E}_{I^{GT}}[log(1-\mathcal{D}^{B}(I^{GT}))]}\\+{\mathbb{E}_{I^{B}}[log(\mathcal{D}^{B}(I^{B}))]},
	\end{aligned}
	\label{eq:ldiss}
\end{equation}
\textbf{Base Adversarial Loss}: We employ the adversarial loss to constrain the realism of our generated images:
\begin{equation}
	\begin{aligned}
		\mathcal{L}^{B}_{a}={\mathbb{E}_{I^{B}}[log(1-\mathcal{D}^{B}(I^{B}))]},
	\end{aligned}
	\label{eq:lgan}
\end{equation}
\textbf{Base Reconstruction Loss}: We achieve visual reconstruction by constraining the l1 loss between the generated base face and the Ground Truth:
\begin{equation}
	\begin{aligned}
		\mathcal{L}^{B}_{r}=\frac{1}{N}\sum_{i=1}^{N}||I^{B}-I^{GT}||_{1},
	\end{aligned}
	\label{eq:lrecon}
\end{equation}
\textbf{Base LPIPS Loss}: We employ the Learned Perceptual Image Patch Similarity loss~\cite{zhang2018unreasonable} to constrain the generated images:
\begin{equation}
	\begin{aligned}
		\mathcal{L}^{B}_{l}=\frac{1}{N}\sum_{i=1}^{N}LPIPS(I^{B},I^{GT}),
	\end{aligned}
	\label{eq:LPIPSloss}
\end{equation}
\textbf{Base Audio-Visual Sync Loss}: We follow~\cite{park2022synctalkface} use the audio-visual sync module proposed in ~\cite{chung2016out,prajwal2020lip}. We train the audio-visual sync module, $\mathcal{F}^{A}$ and $\mathcal{F}^{V}$, on LRS2 \cite{afouras2018deep} datasets and no fine-tune on any generated frames. The generated 5 frames (lower half only) correspond to one audio segment, and the features obtained by $\mathcal{F}^{A}$ and $\mathcal{F}^{V}$ are represented as $f_{a}$ and $f_{v}$, respectively. The outputs features' binary cross-entropy of cosine similarity is computed as follows:
\begin{equation}
	\begin{aligned}
		d_{sync}(f_{a},f_{v})=\frac{f_{a} \cdot f_{v}}{{||f_{a}||_{2}} \cdot {||f_{v}||_{2}}},
	\end{aligned}
	\label{eq:Audio-VisualSyncdiss}
\end{equation}

The Audio-Visual Sync Loss can be formulated as:
\begin{equation}
	\begin{aligned}
		\mathcal{L}^{B}_{av}=-\frac{1}{N}\sum_{i=1}^{N}(log(d_{sync}(\mathcal{F}^{A}(A^{M}_{i}),\mathcal{F}^{V}(\mathbf{I}^{B}_{i})))),
	\end{aligned}
	\label{eq:Audio-VisualSyncloss}
\end{equation}
where $\mathbf{I}^{B}_{i}=\{I^{B}_{n}\}^{i+2}_{n=i-2}$.

To summarize, the training loss for the base face generation stage can be formulated as follows:
\begin{equation}
	\begin{aligned}
		\mathcal{L}^{B}_{total}=\lambda_{a}^{B}\mathcal{L}^{B}_{a}+\lambda_{r}^{B}\mathcal{L}^{B}_{r}+\lambda_{l}^{B}\mathcal{L}^{B}_{l}+\lambda_{av}^{B}\mathcal{L}^{B}_{av},
	\end{aligned}
	\label{eq:basetotalloss}
\end{equation}
where $\lambda_{a}^{B}$, $\lambda_{r}^{B}$, $\lambda_{l}^{B}$, $\lambda_{av}^{B}$ are the hyper-parameter weights.
\subsection{High-Fidelity Rendering}
\subsubsection{HRDecoder}
We have constructed a relatively simple High-Resolution Decoder (HRDecoder) consisting of a base convolution module, an upsampling convolution module, and an output convolution block. The transposed convolution in the upsampling convolution module can convert lower resolution features to higher resolution features.
HRDecoder takes the concatenation of the base face generated in the first stage and the corresponding face landmark sketch as input and outputs a high-fidelity face through the guidance of the landmark sketch.
The high-fidelity rendering process can be formulated as follows:
\begin{equation}
	\begin{aligned}
		I^{HR}=\mathcal{F}^{HR}(I^{B}\oplus I^{S}),
	\end{aligned}
	\label{eq:hr}
\end{equation}
where, $\oplus$ indicate concatenation, $I^{S}$ is the face landmark sketch, $I^{HR}$ is the high-fidelity face, and $\mathcal{F}^{HR}$ is the HRDecoder.
We utilize the mediapipe tool~\cite{lugaresi2019mediapipe} to detect the face landmark sketch from the base face.
To optimize the HRDecoder, we use the model trained in the first stage to generate corresponding base faces and landmark sketches on the dataset as the training dataset for this stage.
\subsubsection{Loss Function for High-Fidelity Rendering}
To get high-fidelity faces, we define a discriminator, HRDiscriminator, at this stage.  We use $\mathcal{D}^{HR}$ to denote the HRDiscriminator. We train HRDiscriminator by adding the following loss:
\begin{equation}
	\begin{aligned}
		\mathcal{L}^{HR}_{disc}={\mathbb{E}_{I^{GT}}[log(1-\mathcal{D}^{HR}(I^{GT}))]}\\+{\mathbb{E}_{I^{HR}}[log(\mathcal{D}^{HR}(I^{HR}))]},
	\end{aligned}
	\label{eq:ldisshr}
\end{equation}
\textbf{HR Adversarial Loss}: Same as Eq.~\ref{eq:lgan}, we employ adversarial loss to constrain the realism of HRDecoder:
\begin{equation}
	\begin{aligned}
		\mathcal{L}^{HR}_{a}={\mathbb{E}_{I^{HR}}[log(1-\mathcal{D}^{HR}(I^{HR}))]},
	\end{aligned}
	\label{eq:lganhr}
\end{equation}
\textbf{HR Perceptual Loss}: We employ the pre-trained VGG~\cite{simonyan2014very}, indicated as $\phi$,  to extract the image features and caculate the features l1 loss to constrain the generated images:
\begin{equation}
	\begin{aligned}
		\mathcal{L}^{HR}_{p}=\frac{1}{N}\sum_{i=1}^{N}||\phi(I^{HR})-\phi(I^{GT})||_{1},
	\end{aligned}
	\label{eq:lreconPerceptual}
\end{equation}
\textbf{HR Reconstruction Loss}: We also employ same as Eq.~\ref{eq:lrecon}, constraining the l1 loss between the generated high-fidelity face and the GT:
\begin{equation}
	\begin{aligned}
		\mathcal{L}^{HR}_{r}=\frac{1}{N}\sum_{i=1}^{N}||I^{HR}-I^{GT}||_{1},
	\end{aligned}
	\label{eq:Reconstructionlosshr}
\end{equation}
\textbf{HR Lip Loss}: To better optimize the lip region, we used the mask of the lip region to constrain the loss of lpips and the reconstruction loss of the lip region.
\begin{equation}
	\begin{aligned}
		\mathcal{L}^{HR}_{l}=\frac{1}{N}\sum_{i=1}^{N}(LPIPS(I^{HR}_{lip},I^{GT}_{lip})\\+||(I^{HR}-I^{GT})*I_{lip}^{mask}||_{1}),
	\end{aligned}
	\label{eq:liploss}
\end{equation}
where $I^{HR}_{lip}$ and $I^{GT}_{lip}$ are the corped according to the lip bounding box of $I^{HR}$ and $I^{GT}$, $I_{lip}^{mask}$ is the lip mask.

The training loss for the high-fidelity rendering is:
\begin{equation}
	\begin{aligned}
		\mathcal{L}^{HR}_{total}=\lambda_{a}^{HR}\mathcal{L}^{HR}_{a}+\lambda_{p}^{HR}\mathcal{L}^{HR}_{p}\\+\lambda_{r}^{HR}\mathcal{L}^{HR}_{r}+\lambda_{l}^{HR}\mathcal{L}^{HR}_{l},
	\end{aligned}
	\label{eq:lav}
\end{equation}
where $\lambda_{a}^{HR}$, $\lambda_{p}^{HR}$, $\lambda_{r}^{HR}$, $\lambda_{l}^{HR}$ are the hyper-parameter weights.
\section{Experiments}

\subsection{Experimental Settings}
\textbf{Implementation Details.}  
We follow~\cite{prajwal2020lip,park2022synctalkface} to process video frames with the centered crops of size $128\times128$ at $25$ fps, and calculate Mel-spectrograms of size $16\times80$ from 16kHz audios using a window size of 800 and hop size of 200. 
For HyperLips-HR, we set the output upsampling to HR$\times 1$ by default, i.e. no upsampling, and the size of the output image remains at $128\times128$.
Hyper-parameters are empirically set: $\lambda_{a}^{B}$=0.2, $\lambda_{r}^{B}$=0.5, $\lambda_{l}^{B}$=0.5, $\lambda_{av}^{B}$=0.3, $\lambda_{a}^{HR}$, $\lambda_{p}^{HR}$, $\lambda_{r}^{HR}$, $\lambda_{l}^{HR}$ are all set to 1.
When training the HyperLips-Base and HyperLips-HR models, we set the learning rate to 0.0001 and used the Adam optimizer in PyTorch. All experiments are performed on a single NVIDIA TITAN RTX GPU.
\vspace{3mm}

\textbf{Dataset.} Two audio-visual datasets, LRS2~\cite{Afouras18c} and MEAD-Neutral~\cite{wang2020mead}, are used in our experiments. 
\textbf{LRS2} is a sentence-level dataset with over 140,000 utterances, consists of 48,164 video clips from outdoor shows on BBC television. We randomly sample 80 videos from the test set for evaluating algorithms quantitatively. 
\textbf{MEAD-Neutral} is a part of MEAD dataset. MEAD dataset records around 40 hours emotional in-the-lab videos at 1080P resolution. We select a total of 1610 videos with neutral emotion and frontal view as MEAD-Neutral dataset and another 80 videos for testing.


\begin{table}[t]
	\fontsize{8}{10} \selectfont
	\centering
	\caption{Quantitative comparison with state-of-the-art talking face generation methods on LRS2 \cite{afouras2018deep} datasets. $\uparrow$ indicates higher is better while $\downarrow$ indicates lower is better.}
\label{tab:table_lrs2}
	\setlength{\tabcolsep}{0.2mm}{
		\begin{tabular}{c|ccccccc|c}
			\hline
			\textbf{Method}&                                   \textbf{PSNR$\uparrow$} & \textbf{SSIM$\uparrow$}  & \textbf{LMD$\downarrow$} & \textbf{LSE-C$\uparrow$}  & \textbf{LSE-D$\downarrow$}  \\ \hline
			
			Wav2Lip \cite{prajwal2020lip}   &    31.794  & 0.894          &1.471          & \textbf{6.841}          & 7.202               \\
			ATVGnet \cite{chen2019hierarchical}        &   32.812  & 0.871          & 1.984         & 4.610         & 8.445               \\
			
			SyncTalkFace \cite{park2022synctalkface}  &   32.138 & 0.886          & 1.354          &4.725          & 8.368               \\

			IP\_LAP \cite{zhong2023identity}                       &       33.281                            & 0.891          &1.494          &3.435          & 9.398      \\

			\textbf{HyperLips-Base(Ours)}                    &  {33.953} & {0.914} & \textbf{1.186} & {6.707} & \textbf{6.878}            \\ 
			\textbf{HyperLips-HR(Ours)}                    &  \textbf{34.914} & \textbf{0.820} & {1.203} & {5.939} & {7.504}            \\ 
			
			\hline
			Ground Truth & N/A& 1.000& 0.000& 8.354& 6.204 \\
			\hline
		\end{tabular}%
	}

\end{table}

\begin{table}[t]
	\fontsize{8}{10} \selectfont
	\centering
	\caption{Quantitative comparison with state-of-the-art talking face generation methods on MEAD-Neutral \cite{wang2020mead} datasets. $\uparrow$ indicates higher is better while $\downarrow$ indicates lower is better. }
\label{tab:table_mead}
	\setlength{\tabcolsep}{0.2mm}{
		\begin{tabular}{c|ccccccc|c}
			\hline
			\textbf{Method}&                                   \textbf{PSNR$\uparrow$} & \textbf{SSIM$\uparrow$}  & \textbf{LMD$\downarrow$} & \textbf{LSE-C$\uparrow$}  & \textbf{LSE-D$\downarrow$}  \\ \hline
			
			Wav2Lip \cite{prajwal2020lip}   &    29.867  & 0.683          &2.294          &\textbf{2.312}          & \textbf{10.488}               \\

			DINet(O)\footnotetext[1]{hello}\cite{zhang2023dinet}  &   30.056 & 0.707         & 2.367          &{1.710}         & {11.560}               \\
			DINet(R)\cite{zhang2023dinet}  &   28.573 & 0.621         & 1.838          &1.203        & 11.906              \\
			
			IP\_LAP  \cite{zhong2023identity}       &   30.578  & 0.699          & 1.386        &  {1.349}         & 11.868               \\
			\textbf{HyperLips-Base(Ours) }                  &  {30.784} & {0.721} & {1.294} & {1.266} & {11.854}            \\ 
			\textbf{HyperLips-HR(Ours) }	                &  \textbf{31.503} & \textbf{0.747} & \textbf{1.237} & {1.271} & {11.861}            \\ 
			\hline
			Ground Truth & N/A& 1.000& 0.000& 2.161& 11.091 \\
			\hline
		\end{tabular}%
		
	}

\end{table}

\begin{figure*}[ht]
	\vspace{-5pt}
	\centering
	\includegraphics[width=1.0\linewidth]{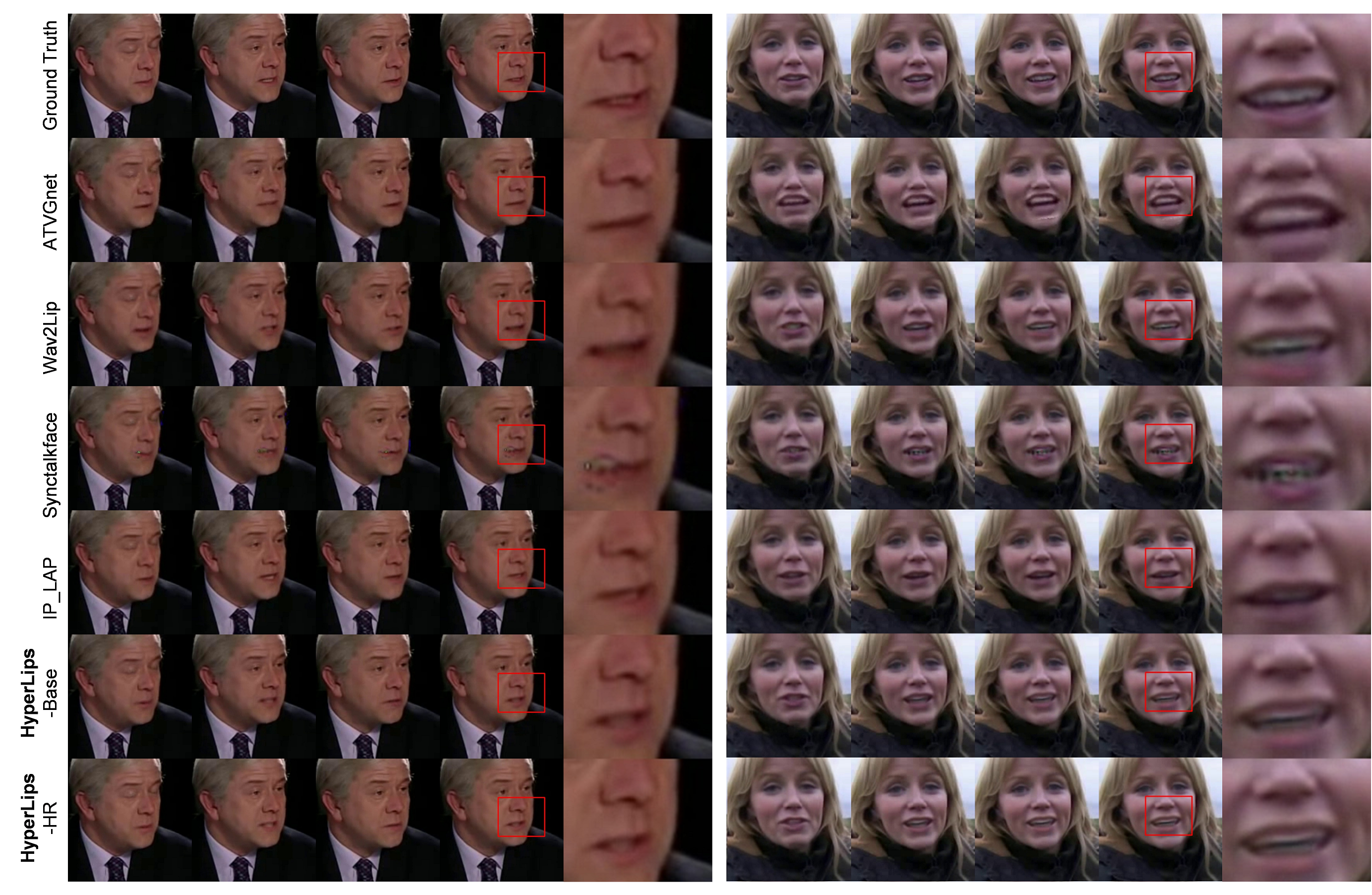}
	\vspace{-18pt}
	\caption{Qualitative comparisons with state-of-the-art methods on LRS2 datasets. Our method is capable of rendering more {high-fidelity} faces. More results are presented in the supplementary material.
	}
	\vspace{-8pt}
	\label{fig:Qualitativelrs2}
\end{figure*}
\begin{figure*}[ht]
	\vspace{-5pt}
	\centering
	\includegraphics[width=1.0\linewidth]{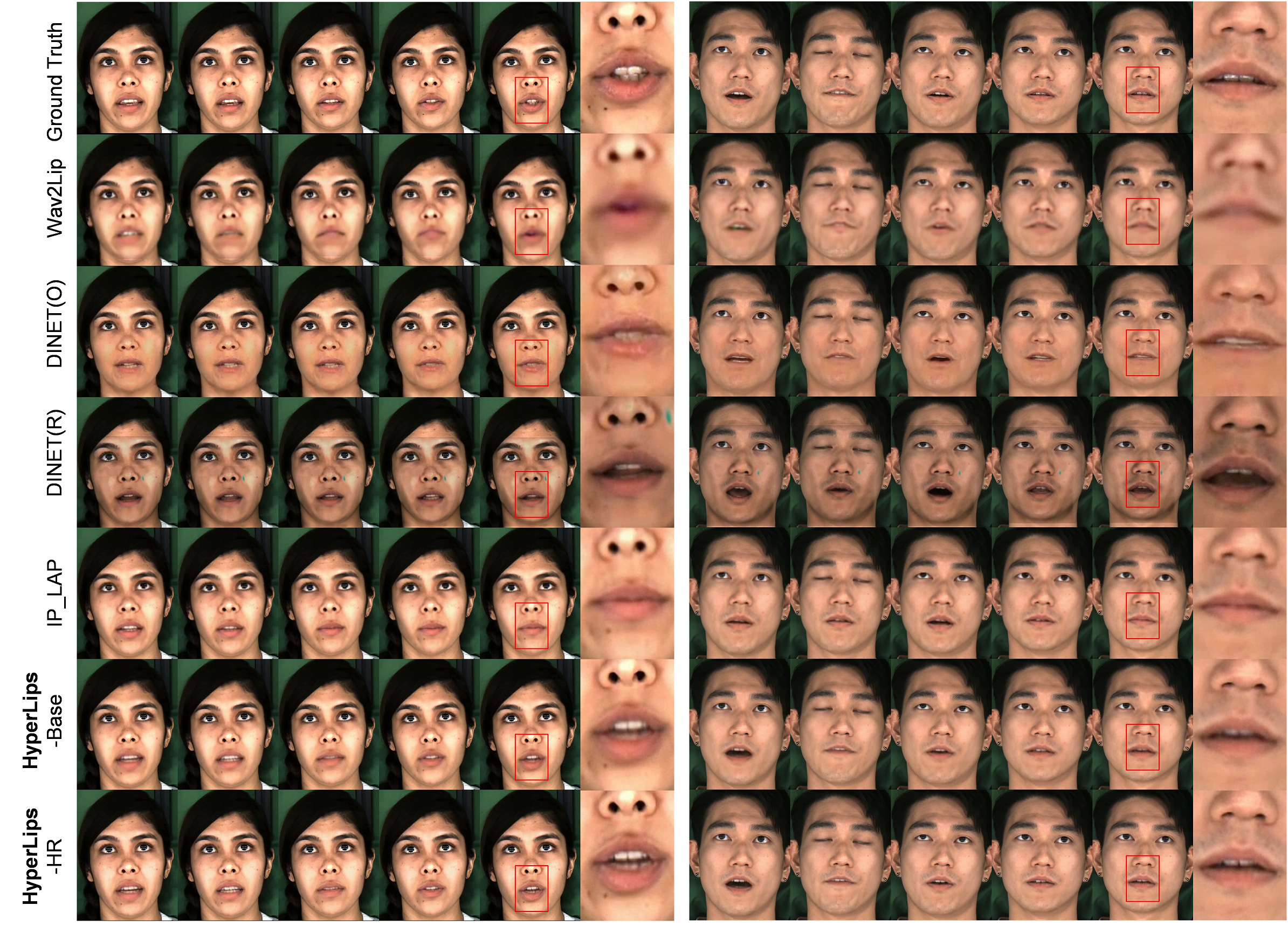}
	\vspace{-18pt}
	\caption{Qualitative comparisons with state-of-the-art methods on MEAD-Neutral datasets. Our method is capable of rendering more {high-fidelity} faces. More results are presented in the supplementary material.
	}
	\vspace{-8pt}
	\label{fig:Qualitativemead}
\end{figure*}
\textbf{Comparison Methods.}
We compare our method against state-of-the-art methods~\cite{prajwal2020lip,chen2019hierarchical,park2022synctalkface,zhong2023identity,zhang2023dinet} on the person-generic audio-driven talking face generation. 
\textbf{Wav2Lip}~\cite{prajwal2020lip} uses an encoder-decoder model learned via adversarial training to produce talking face videos.
\textbf{ATVGnet}~\cite{chen2019hierarchical} takes advantage of 2D landmarks to generate talking face videos from the input audio and an identity frame. 
\textbf{SyncTalkFace}~\cite{park2022synctalkface} proposes Audio-Lip Memory that brings in visual information of the mouth region corresponding to input audio and enforces fine-grained audio-visual coherence. 
\textbf{IP\_LAP}~\cite{zhong2023identity} proposes a two-stage framework consisting of audio-to-landmark generation and landmark-to-video rendering procedures.
\textbf{DINet}~\cite{zhang2023dinet} proposes a Deformation Inpainting Network for high-resolution face visually dubbing. 
For more comparison settings, please refer to our supplementary document.

\subsection{Evaluation Metrics}
We use Peak Signal-to-Noise Ratio (\textbf{PSNR}) and Structured similarity (\textbf{SSIM}) \cite{wang2004image} to measure the similarity between generated and ground-truth images. And we use dlib \cite{king2009dlib} to detect the lip landmark distances (\textbf{LMD}) between ground truth frames and those of generated frames. \textbf{LSE-C} and \textbf{LSE-D} proposed by \cite{prajwal2020lip} are cibfudebce score (higher the better) and distance score (lower the better) between audio and video features from SyncNet \cite{chung2016out}, respectively. LSE-C and LSE-D measure correspondence between audio and visual features while LMD directly measures visual to visual coherence. For a fair comparison, we evaluate the cropped region of the face based on the face detector used in {Wav2Lip}~\cite{prajwal2020lip}.

We generate corresponding videos using different methods based on different audio in the test dataset. Specifically, the face in the video frame is first detected by face detection. Then, the corresponding face area is resized according to the required resolution size of the corresponding method. After the face is generated by the corresponding method, it is pasted back into the original video.
For a fair comparison, frames extracted from talking face videos, which are cropped based on the face detector used in Wav2Lip are resized to $160\times 160$.
When calculating the related metrics, we detect faces in the generated video and the corresponding ground truth video, resize them to $160\times 160$, and then perform frame-by-frame calculations.
Wav2Lip synthesizes face with $96\times 96$ resolution; DINet synthesizes face with $416\times 320$ resolution; ATVGnet, IP\_LAP, SyncTalkFace, and ours synthesize face with $128\times 128$ resolution.
For LSE-D and LSE-C, we generate talking face videos by inputting audio and face come from the different videos in test datasets and use SyncNet to calculate LSE-C and LSE-D with generating talking face videos.

\subsection{Quantitative Comparison}
Table~\ref{tab:table_lrs2} and ~\ref{tab:table_mead} show the quantitative comparison on the LRS2 and MEAD-Neutral datasets, respectively. DINet(O) indicates tested on the MEAD dataset using the checkpoints officially released by DINet. DINet(R) is the result of our reproduction on the MEAD-Neutral dataset according to the code of DINet. In the tables, our HyperLips-HR output resolution is $128 \times 128$ without upsampling.
The results show that whether it is our HyperLips-Base or our HyperLips-HR, the generated faces are significantly better than other methods in terms of PSNR, SSIM, and LMD metrics. 
Our HyperLips-HR is significantly better than our HyperLips-Base in terms of PSNR and SSIM, which shows that our HRDecoder has enhanced high-fidelity face rendering. However, there is no significant increase in the LMD metric, which shows that HRDecoder does not help improve lip synchronization.
Regarding PSNR and SSIM, our results in Table 1 are better than those in Table 2. This is because the face quality in the LRS2 dataset is worse than that in the MEAD dataset, making it easier for the faces generated by our model to reach the quality of LRS2.

For LSE-C and LSE-D, Wav2Lip perform better results and even outperforms those of ground truth. The weights~\footnote[1]{SyncNet Weights:  http://www.robots.ox.ac.uk/~vgg/software/lipsync/data/\\syncnet\_v2.model  } of SyncNet we used in the test were derived from~\cite{chung2016out} without fine-tuning. In fact, these two metrics have been discussed in ~\cite{zhou2021pose,park2022synctalkface}, it only proves that their lip-sync results are nearly comparable to the ground truth, not better. 
On the one hand, the dataset used for model training may not match the distribution of the dataset we tested, resulting in two test results that may not accurately reflect lip synchronization; on the other hand, we performed better on the LMD metric, which is another synchronization metric that measures correspondence in the visual domain.

\subsection{Qualitative Comparison}

\begin{figure}[ht]
	\vspace{-4pt}
	\centering
	\includegraphics[width=0.8\linewidth]{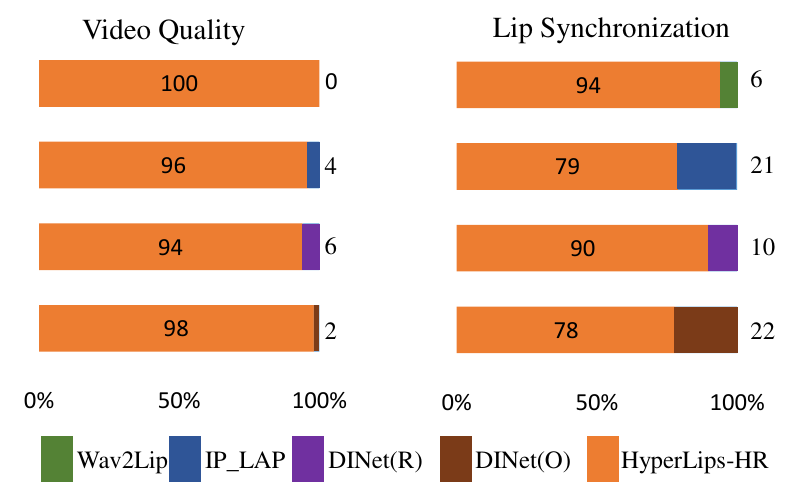}
	\vspace{-1pt}
	\caption{User study about video  quality and lip synchronization.
	}
	\vspace{-8pt}
	\label{fig:userstudy}
\end{figure}

\textbf{User Study.}~To verify the video quality and lip synchronization of our talking face generation method, we invited 20 participants to evaluate the generated videos. We randomly selected 5 videos from the MEAD-Neutral~\cite{wang2020mead} test dataset and generated different videos using different methods: {Wav2Lip}~\cite{prajwal2020lip}, {IP\_LAP}~\cite{zhong2023identity}, {DINet(R)}~\cite{zhang2023dinet}, {DINet(O)}~\cite{zhang2023dinet} and HyperLips-HR(Ours). We asked the participants to vote for the video in two evaluation indicators: video quality of the results and whether to keep the lip synchronization. We collected 100 votes for each evaluation indicator and presented the result as a box plot in Fig. ~\ref{fig:userstudy}. As can be seen, our results stand out from other methods in terms of video quality and lip synchronization.

\textbf{Visualization Comparison.} 
Fig.~\ref{fig:Qualitativelrs2} and Fig.~\ref{fig:Qualitativemead} show examples from LRS2 and MEAD-Neural dataset, respectively. Compared to other methods, our method produces images that are visually closer to the ground truth and show no artifacts in our results. The superiority of our method cannot be seen on the LRS2 dataset because the faces in this dataset are relatively blurred. But on the MEAD dataset, our method produces results that render faces clearly, and even teeth can be seen clearly.
Our method also excels in lip sync.
 For example, in the last face on the left in Fig.~\ref{fig:Qualitativemead}, our results perfectly reproduce the current mouth shape, which is slightly open with teeth exposed, but the results from IP\_LAP are not.

\begin{table}[!t]
	\fontsize{8}{10} \selectfont
	\centering
	\caption{Ablation study on the size of HRDecoder output.}
\label{tab:tablesizeHRDecoderoutput}
	\setlength{\tabcolsep}{0.1mm}{
		\begin{tabular}{c|c|ccccccc}
			\hline
			\textbf{Dataset}& \textbf{Size}&                                  \textbf{PSNR$\uparrow$} & \textbf{SSIM$\uparrow$}  & \textbf{LMD$\downarrow$} & \textbf{LSE-C$\uparrow$}  & \textbf{LSE-D$\downarrow$}  \\ \hline
			\multirow{3}{*}{LRS2}  & S=128(Base) & {33.953} & {0.914} & {1.186} & \textbf{6.707} &\textbf{6.878} \\
			& S=128(HR$\times$1)                    &  {34.914} & {0.920} & {1.203} & {5.939} & {7.504}            \\ 
			&S=256(HR$\times$2)                     &  \textbf{35.159} & \textbf{0.924} & \textbf{1.070} & {4.965} & {8.400}            \\ 
			&S=512(HR$\times$4)                      &  {34.895} & {0.921} & {1.175} & {5.580} & {7.835}            \\ 
			\hline
			\multirow{3}{*}{MEAD-Neural}&S=128(Base) &  {30.784} & {0.721} & {1.294} & {1.266} & \textbf{11.854}            \\ 
			& S=128(HR$\times$1)   &  {31.503} & {0.747} & {1.237} & \textbf{1.271} & {11.861}            \\ 
			&S=256(HR$\times$2)    &  \textbf{31.539} & {0.747} & \textbf{1.204} & {1.265} & {11.863}            \\ 
			&S=512(HR$\times$4)   &  {31.460} & \textbf{0.750} & {1.210} & {1.255} & {11.879}            \\ 
			\hline
		\end{tabular}%
		
	}

\end{table}

\begin{table}[!t]
	\fontsize{8}{10} \selectfont
	\centering
	\caption{Ablation study on effectiveness for sketch input of HRDecoder.}
\label{tab:tablesketchHRDecoder}
	\setlength{\tabcolsep}{1.0mm}{
		\begin{tabular}{c|c|cccccc|c}
			\hline
			\textbf{Method}&                                   \textbf{PSNR$\uparrow$} & \textbf{SSIM$\uparrow$}  & \textbf{LMD$\downarrow$} & \textbf{LSE-C$\uparrow$}  & \textbf{LSE-D$\downarrow$}  \\ \hline
			
			Base                   &  {30.784} & {0.721} & {1.294} & {1.266} &{11.854}            \\ 
			\hline	
			
			HR$\times$1(w/o sketch)     &   30.762 & 0.727         & 1.305          &1.244         &{11.938}               \\
			HR$\times$1(w/\;\; sketch)      &  \textbf{31.503} & \textbf{0.747} & \textbf{1.237} & \textbf{1.271} & \textbf{11.861}            \\ 
			\hline	
			HR$\times$2(w/o sketch)     &   30.543 & 0.734         & 1.300          &1.225        & \textbf{11.853}              \\
			HR$\times$2(w/\;\; sketch)      & \textbf {31.539} & \textbf{0.747} & \textbf{1.204} & \textbf{1.265} & {11.863}            \\ 
			\hline	
			HR$\times$4(w/o sketch)     &   30.281  & 0.730          & 1.348        &  {1.216}         & 12.004               \\			
			HR$\times$4(w/\;\; sketch)      & \textbf{31.460} &\textbf{0.750} & \textbf{1.210} & \textbf{1.255} & \textbf{11.879}            \\ 
			\hline

		\end{tabular}%
		
	}

\end{table}

\begin{table}[t]
	\fontsize{8}{10} \selectfont
	\centering
	\caption{Ablation study on effectiveness for Finetuning.}
\label{tab:AblationFineturn}
	\setlength{\tabcolsep}{0.3mm}{
		\begin{tabular}{c|c|cccccc|c}
			\hline
			\textbf{Method}&                                   \textbf{PSNR$\uparrow$} & \textbf{SSIM$\uparrow$}  & \textbf{LMD$\downarrow$} & \textbf{LSE-C$\uparrow$}  & \textbf{LSE-D$\downarrow$}  \\ \hline

			Base(w/o fineturn, MEAD)     &   31.200 & 0.834         & 1.259          &3.050         &{10.463}               \\
			Base(w/\;\; fineturn, MEAD)     &   31.672 & \textbf{0.850}         & 1.199          &3.210         &{10.527}               \\
			HR(w/\;\; fineturn, MEAD)      &  \textbf{31.917} & {0.842} & \textbf{1.164} & \textbf{3.594} & \textbf{10.239}            \\ 
			\hline
			Base(w/o fineturn, LRS2)     &   31.303 & 0.826         & \textbf{1.150}          &\textbf{4.054}         &\textbf{9.587}               \\
			Base(w/\;\; fineturn, LRS2)     &   31.846 & \textbf{0.851}        & 1.164          &3.516         &{10.565}               \\
			HR(w/\;\; fineturn, LRS2)      &  \textbf{32.012} & {0.845} & {1.164} & {3.516} & {10.261}            \\ 
			\hline
			GT     &  N/A & {1.000} & {0.000} & {3.615} & {10.309}            \\ 
			\hline

		\end{tabular}%
		
		\vspace{-5pt}
	}

	\vspace{-5pt}
\end{table}

\subsection{Ablation Study}
In this section, we perform ablation studies to validate the effect of core components in our method and the performance gain derived from high-fidelity rendering.

\textbf{The Size of HRDecoder Output.}
Our input in the high-fidelity rendering stage is fixed at $128 \times 128$, and the output can render faces with different resolutions, such as $128 \times 128$ (HR $\times$ 1), $256 \times 256$ (HR $\times$ 2), and $512 \times 512$ (HR $\times$ 4), through the transposed convolution of HRDecoder.
High-resolution faces can often produce finer images, which are convenient for application to high-resolution videos.
In Table~\ref{tab:tablesizeHRDecoderoutput}, we study the effect of different output sizes on the LRS2 and MEAD datasets.
Our HR models significantly compares with the Base model regarding image quality (such as PSNR and SSIM metrics), e.g., PSNR, the HR$\times$1 model is 34.914, and the Base model is 33.953. 
However, in terms of lip synchronization (such as LMD indicators), the HR models are only comparable to the Base model.
However, for all HR models, the image quality index does not increase significantly with increasing size, and even the image quality at size $512 \times 512$ is comparable to that at size $128 \times 128$.
It can be concluded that on the LRS2 and MEDA datasets, a size of $256 \times 256$ can already provide a cost-effective result.


\textbf{Effectiveness for Sketch Input of HRDecoder.}
In HRDecoder, we introduced face landmark sketches to guide the generation of high-fidelity face images.
We performed corresponding ablation experiments on the MEAD dataset to verify the influence of sketches as input to HRDecoder on the rendering results.
In Table~\ref{tab:tablesketchHRDecoder}, ``w/o sketch" means that no face landmark sketches are used as input, and ``w/ sketch" means that face landmark sketches are used as input.
The results show that in all HR models, the rendering results of adding face landmark sketches as a guide input are better than the rendering results without adding sketches.
This suggests that sketches are beneficial for generating high-fidelity face images.

\textbf{Effectiveness for Finetuning.} Although our method supports dubbing any face video, it may generally perform poorly on unseen faces.
Therefore, we performed an ablation experiment on this.
We chose a video of Kate's speech and used 4 minutes and 40 seconds of the video as training data and another 18 seconds as test data.
As shown in Table~\ref{tab:AblationFineturn}, we fine-tuned the pre-trained models on the MEAD and LRS2 datasets to Kate's videos and obtained corresponding results.
The results show that the deformed model produces slightly better results in terms of visual quality without a significant improvement in lip synchronization.

\begin{figure}[ht]
	\vspace{-5pt}
	\centering
	\includegraphics[width=0.9\linewidth]{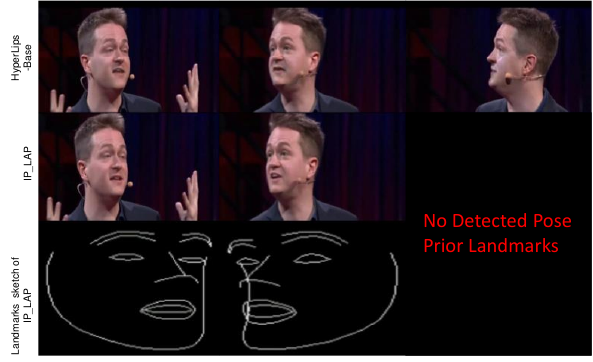}
	\vspace{-5pt}
	\caption{The Impact of Landmark Detection Failure
	}
	\vspace{-8pt}
	\label{fig:ab_sideface}
\end{figure}
\textbf{The Impact of Landmark Detection Failure.} There is a defect in landmark-based talking face generation methods (such as IP\_LAP, ATVGnet, etc.). That is, faces cannot be generated if landmark detection fails. There is no such problem with those that are not based on landmark detection, such as Wav2Lip, etc. Our method is not based on landmark detection in the first stage (Base model), so it can still generate the correct face even if the face deviation is severe, as shown in Fig. ~\ref{fig:ab_sideface}. As for the second stage (HR models), our method still has such shortcomings.

\section{Conclusion}
We propose a hypernetwork for controlling lip movements with audio information to achieve lip synchronization in the task of talking face generation. 
We first use a FaceEncoder to extract the visual face information as latent code from the source video; and then use HyperConv to modify the latent code to synchronize the lip movement with the audio; finally, FaceDecoder will decode the modified and synchronized latent code into visual face content.
The weight parameters are updated by HyperNet using the audio features as input. 
In order to achieve high-fidelity human face rendering, we propose HRDecoder, which uses landmark guidance as the face detail enhancement of faces. Therefore, our method effectively improves lip synchronization and face visual quality.

\bibliographystyle{IEEEtran}
\bibliography{sample-sigconf}
\clearpage

\section{Supplementary Material}
\subsection{Post Processing}
\begin{figure}[ht]
	\vspace{-0pt}
	\centering
	\includegraphics[width=1\linewidth]{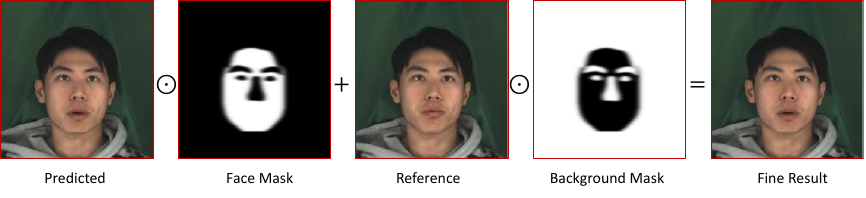}
	\vspace{-0pt}
	\caption{Post Processing for Generated Face Fusion.	}
	\vspace{0pt}
	\label{fig:pastback}
\end{figure}

\textbf{Generated Face Fusion:} In order to obtain a more natural driving effect, we followed~\cite{zhong2023identity} to create a post-processing process. As shown in Fig.~\ref{fig:pastback}, \emph{Predicted} is the face image predicted by our model, denote as $I^{P}$; \emph{Face Mask} is the face mask parsed from the predicted face image using the Face Parsing method, denote as $\alpha$; \emph{Reference}  is the reference face image, denote as $I^{R}$; \emph{Background Mask} is equal to \emph{(1-$\alpha$)}; the \emph{Fine Result} can be formulated as follows: 
\begin{equation}
	\begin{aligned}
		{I}^{fine}=I^{P}\times \alpha+ I^{R}\times(1-\alpha),
	\end{aligned}
	\label{eq:postprocessing}
\end{equation}
We take the generated face image out of the eyebrows, eyes and nose, and paste the rest back to the reference image.

\begin{figure}[ht]
	\vspace{-0pt}
	\centering
	\includegraphics[width=1\linewidth]{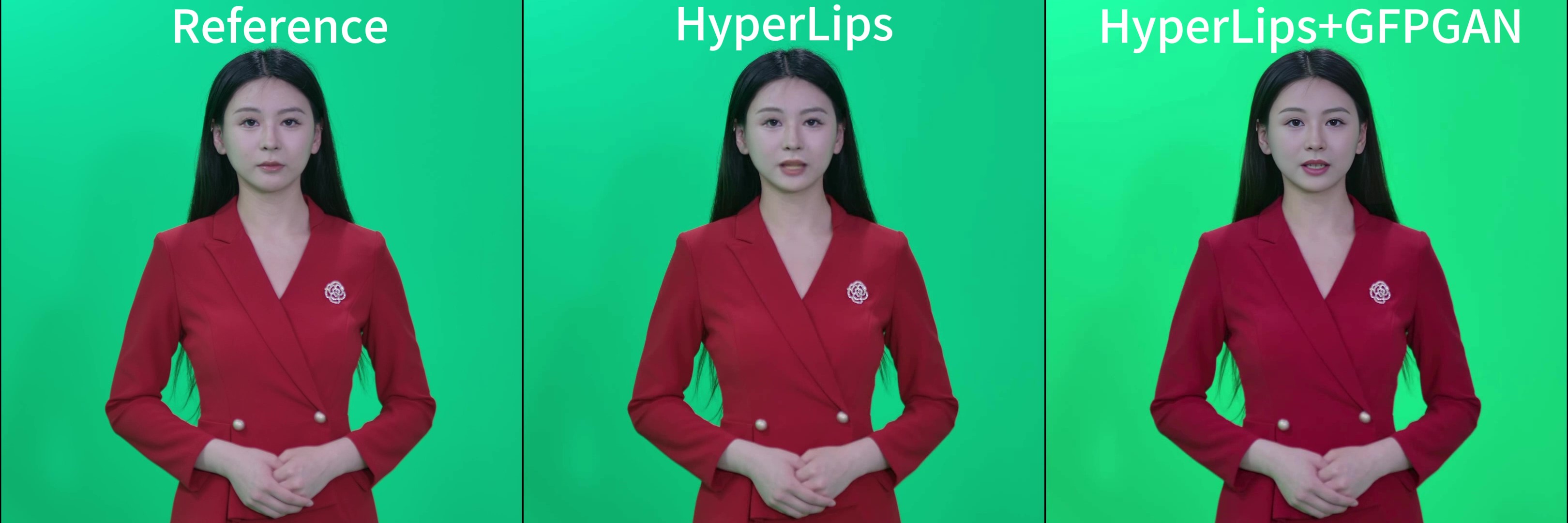}
	\vspace{-0pt}
	\caption{Post Processing for Face Enhencement.	}
	\vspace{0pt}
	\label{fig:FaceEnhencement}
\end{figure}

\textbf{Face Enhencement:} Even though our method can produce a high-fidelity face with clear lips, in the actual production process, it may still not be perfect in the following cases: 1. The reference video (source video) is blurred; 2. Not performed better fine-tuning training. The face restoration algorithm can further improve the quality of the generated faces. Fig.~\ref{fig:FaceEnhencement} shows the enhanced effect of the face restoration algorithm: We drive the reference video with hyperlips to get the lip-sync generated video, and then use GFP-GAN~\cite{wang2021towards} to enhance the generated video. It can be seen that the enhanced video is clearer in facial details. But this method also has obvious disadvantages, that is, after the face is enhanced, there will be artifacts in the enhanced face due to the GAN technology, and there will be a certain deviation from the real original face.

\subsection{Configuration of HyperLips Model}

Our proposed model has been divided into two stages: Base Face Generation and High-Fidelity Rendering. 
Here, we list the configuration of each sub-module included in the two stages in detail

\begin{figure}[ht]
	\vspace{-0pt}
	\centering
	\includegraphics[width=1.0\linewidth]{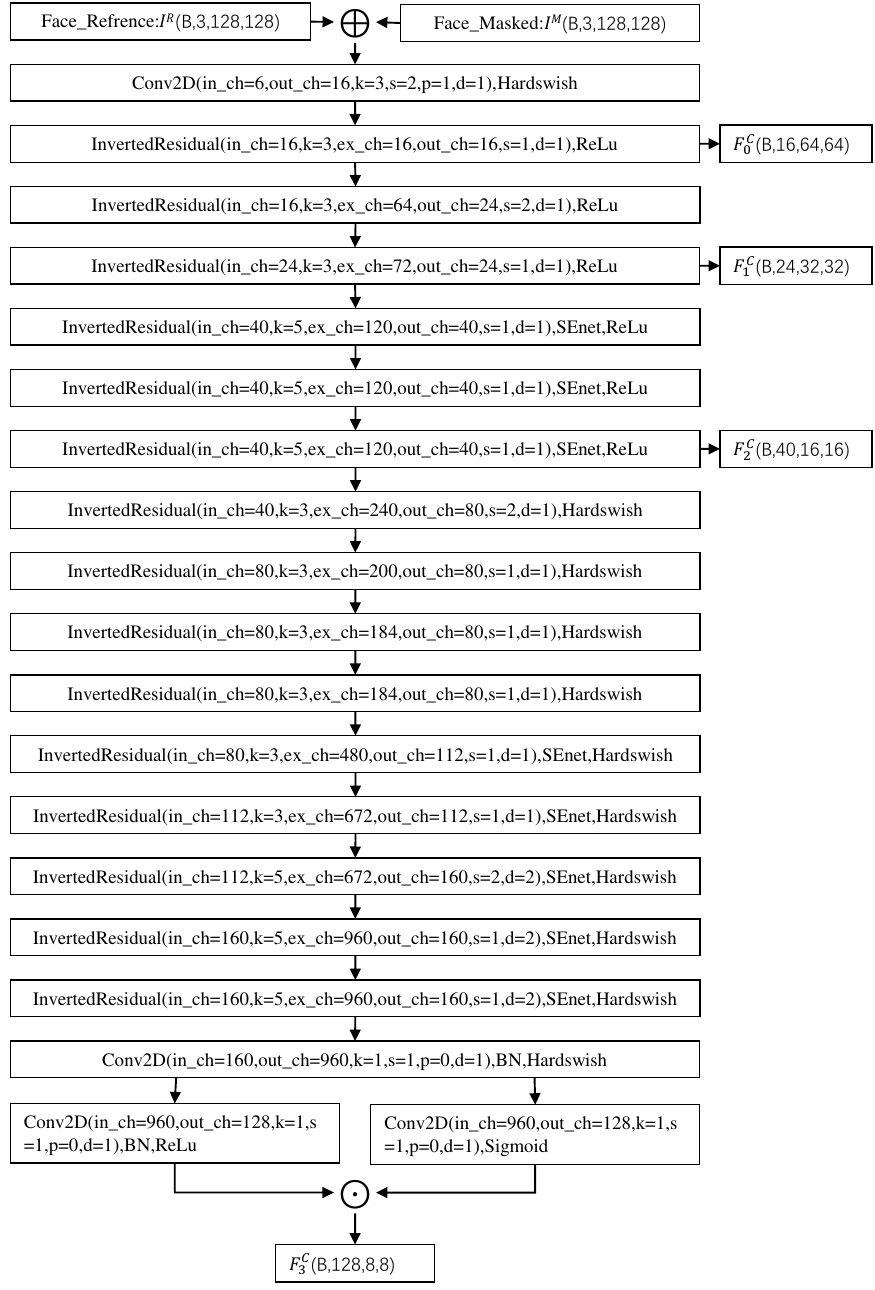}
	\vspace{-0pt}
	\caption{Configuration of FaceEncoder. $\oplus$ indicate concatenation, $\odot$ indicate multiplication, $in\_ch$ indicate the number of input channel, $out\_ch$ indicate the number of out channel, $ex\_ch$ indicate the number of extension channel, $k$ indicate the kernel size, $s$ indicate the stride size, $p$ indicate the padding size. 	}
	\vspace{0pt}
	\label{fig:FaceEncoder}
\end{figure}

\begin{figure}[ht]
	\vspace{-0pt}
	\centering
	\includegraphics[width=1.0\linewidth]{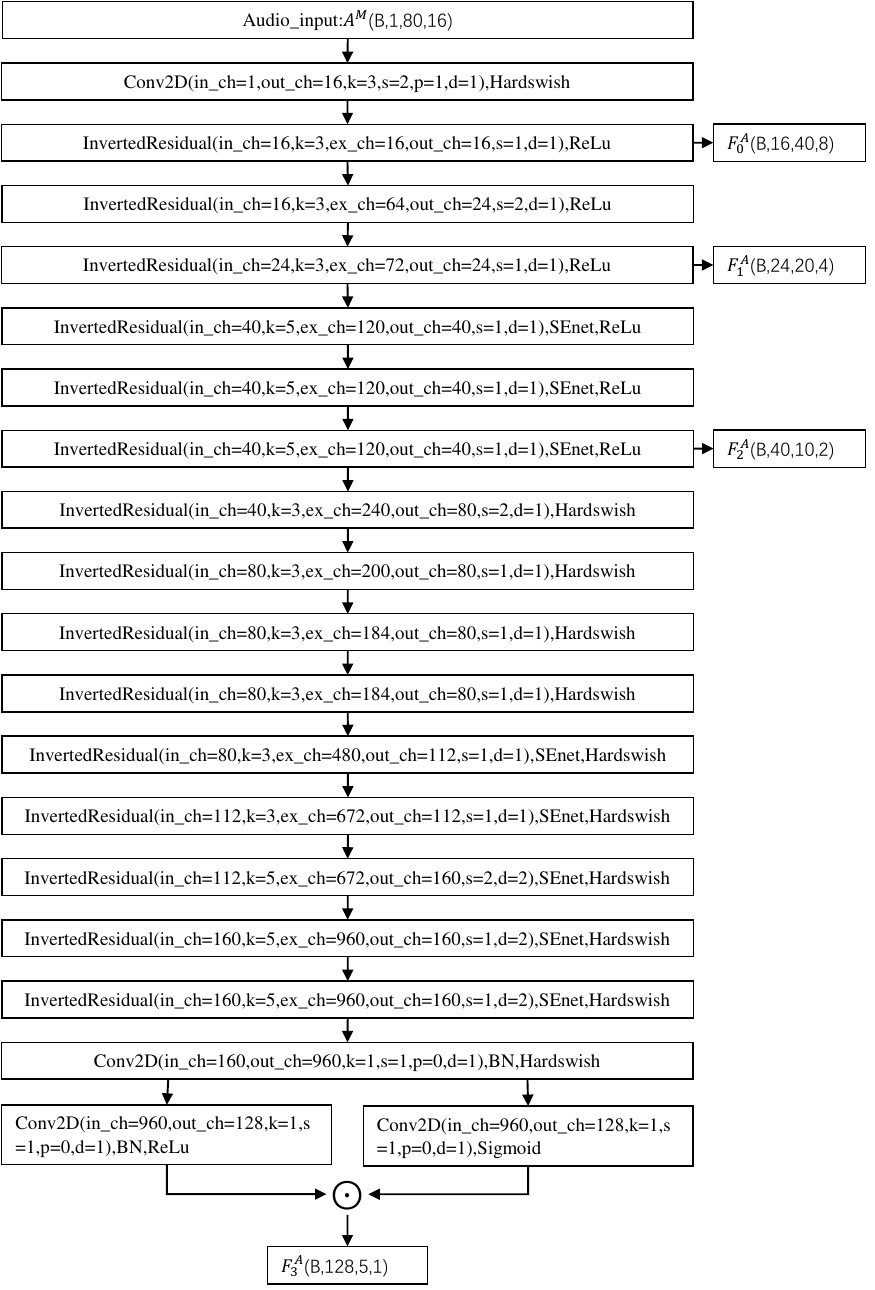}
	\vspace{-0pt}
	\caption{Configuration of AudioEncoder. $\oplus$ indicate concatenation, $\odot$ indicate multiplication, $in\_ch$ indicate the number of input channel, $out\_ch$ indicate the number of out channel, $ex\_ch$ indicate the number of extension channel, $k$ indicate the kernel size, $s$ indicate the stride size, $p$ indicate the padding size. }
	\vspace{0pt}
	\label{fig:AudioEncoder}
\end{figure}

\begin{figure}[ht]
	\vspace{-0pt}
	\centering
	\includegraphics[width=1.0\linewidth]{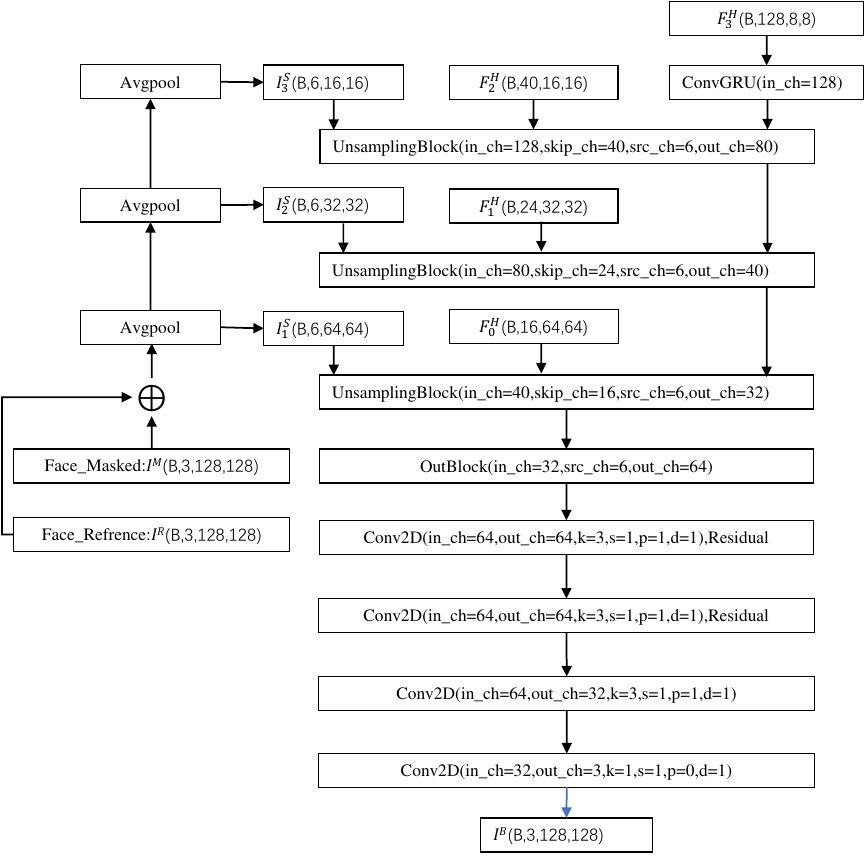}
	\vspace{-0pt}
	\caption{Configuration of FaceDecoder. $\oplus$ indicate concatenation, $in\_ch$ indicate the number of input channel, $out\_ch$ indicate the number of out channel, $k$ indicate the kernel size, $s$ indicate the stride size, $p$ indicate the padding size.	}
	\vspace{0pt}
	\label{fig:FaceDecoder}
\end{figure}

\begin{figure}[ht]
	\vspace{-1pt}
	\centering
	\includegraphics[width=1.0\linewidth]{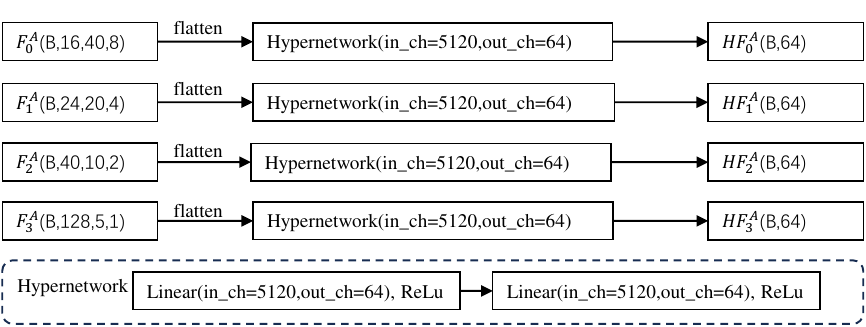}
	\vspace{-0pt}
	\caption{Configuration of HyperNet.	}
	\vspace{0pt}
	\label{fig:HyperNet}
\end{figure}

\begin{figure}[ht]
	\vspace{-0pt}
	\centering
	\includegraphics[width=1.0\linewidth]{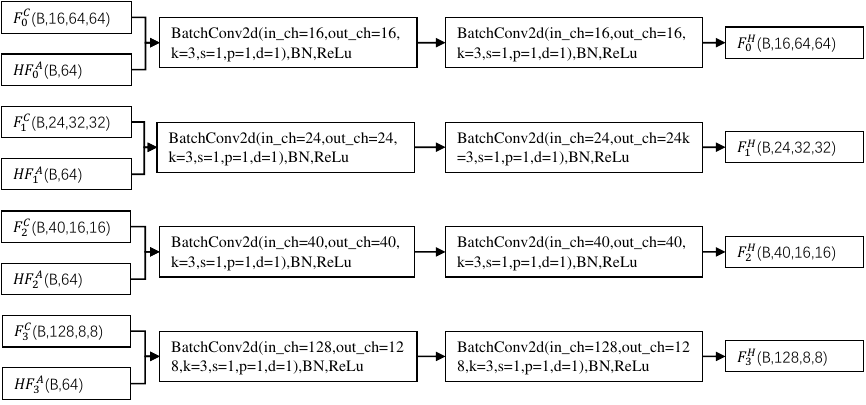}
	\vspace{-0pt}
	\caption{Configuration of HyperConv. $in\_ch$ indicate the number of input channel, $out\_ch$ indicate the number of out channel,  $k$ indicate the kernel size, $s$ indicate the stride size, $p$ indicate the padding size.	}
	\vspace{0pt}
	\label{fig:HyperConv}
\end{figure}

In the Base Face Generation stage, it includes FaceEncoder, AudioEncoder, FaceDecoder, HyperNet, and HyperConv sub-modules. The detailed configuration of these sub-modules is shown in Fig.~\ref{fig:FaceEncoder},~\ref{fig:AudioEncoder},~\ref{fig:FaceDecoder},~\ref{fig:HyperNet},  and ~\ref{fig:HyperConv} respectively.
For FaceEncoder and AudioEncoder, the network architecture is the same, but the input feature size is not the same, therefore, the output feature size is also different.
For FaceEncoder, the input is the concatenation of Refrence and Masked, therefore, the input size is $B\times 6\times128\times128$; For AudioEncoder, the input is Mel-spectrogram of the audio, and the size is $B\times 1\times80\times60$, where $B$ indicate the batch size.

\begin{figure}[ht]
	\vspace{-0pt}
	\centering
	\includegraphics[width=0.6\linewidth]{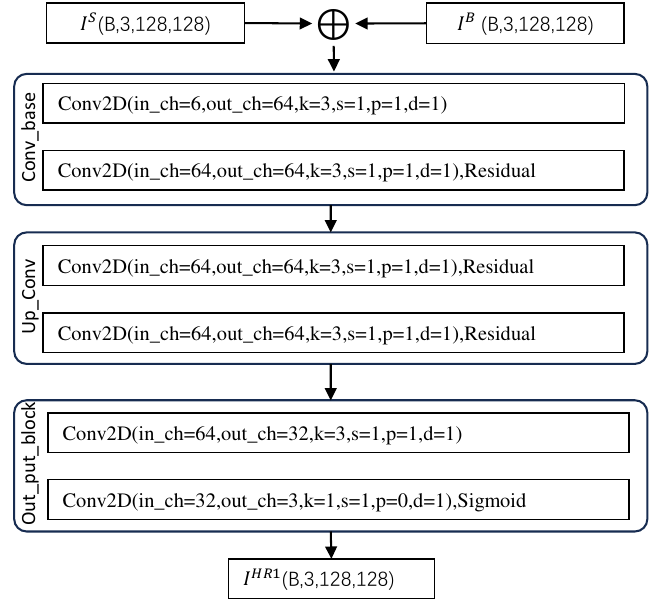}
	\vspace{-0pt}
	\caption{Configuration of HR$\times$1. $\oplus$ indicate concatenation, $in\_ch$ indicate the number of input channel, $out\_ch$ indicate the number of out channel,  $k$ indicate the kernel size, $s$ indicate the stride size, $p$ indicate the padding size. 	}
	\vspace{0pt}
	\label{fig:HR1}
\end{figure}

\begin{figure}[ht]
	\vspace{-0pt}
	\centering
	\includegraphics[width=0.6\linewidth]{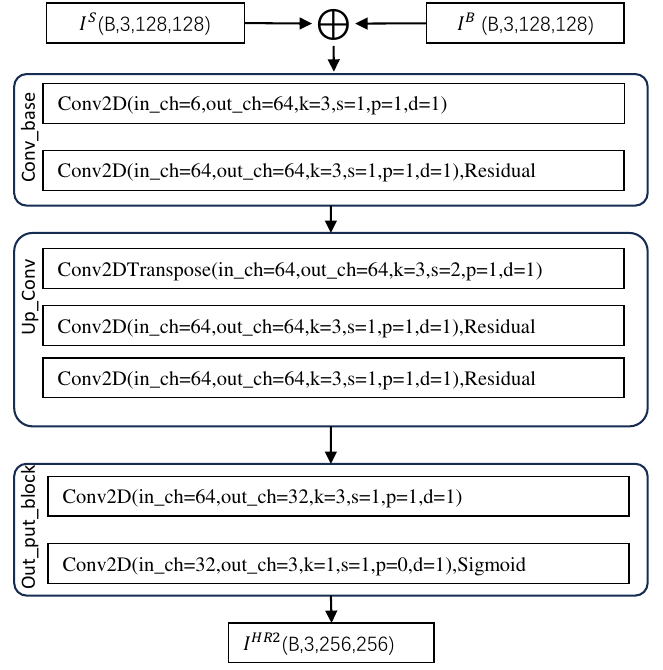}
	\vspace{-0pt}
	\caption{Configuration of HR$\times$2. $\oplus$ indicate concatenation, $in\_ch$ indicate the number of input channel, $out\_ch$ indicate the number of out channel,  $k$ indicate the kernel size, $s$ indicate the stride size, $p$ indicate the padding size.	}
	\vspace{0pt}
	\label{fig:HR2}
\end{figure}

\begin{figure}[ht]
		\vspace{-10pt}
		\centering
		\includegraphics[width=0.6\linewidth]{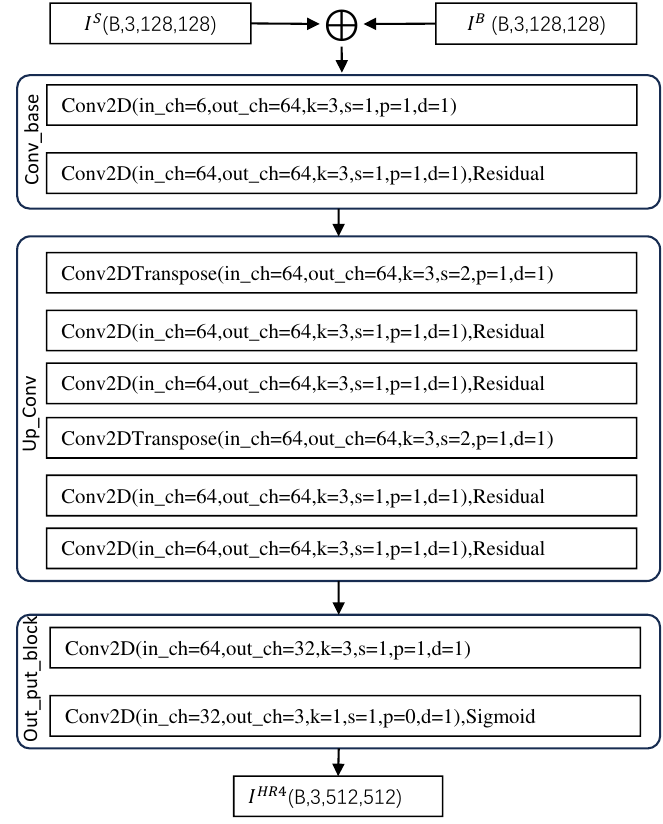}
		\vspace{-0pt}
		\caption{Configuration of HR$\times$4. $\oplus$ indicate concatenation, $in\_ch$ indicate the number of input channel, $out\_ch$ indicate the number of out channel,  $k$ indicate the kernel size, $s$ indicate the stride size, $p$ indicate the padding size.	}
		\vspace{0pt}
		\label{fig:HR4}
\end{figure}

In the High-Fidelity Rendering stage, only one module, HRDecoder, consists of three parts: conv\_base, up\_conv and out\_put\_block. Although HRDecoder is relatively simple, it has three variants that allow the model to render face images of different resolutions: HR$\times$1, the configuration shows in Fig.~\ref{fig:HR1}, and the output image size is 128$\times$128; HR$\times$2, the configuration shows in Fig.~\ref{fig:HR2}, and the output image size is 256$\times$256; HR$\times$4, the configuration shows in Fig.~\ref{fig:HR4}, and the output image size is 512$\times$512.  The input to HRDecoder is the concatenation feature of the base face with the sketch extracted from the base face, and the output is the high-fidelity face.
We use Transpose Convolution to realize the conversion of feature maps from low resolution to high resolution. In HR$\times$1, we did not use transpose convolution, so the output size is the same as the input; in HR$\times$2, we use one transpose convolution, so the output size is twice the input; in HR$\times$4, we use two transpose convolution, so the output size is four times the input.


\clearpage

\begin{figure*}[ht]
	\vspace{-0pt}
	\centering
	\includegraphics[width=1\linewidth]{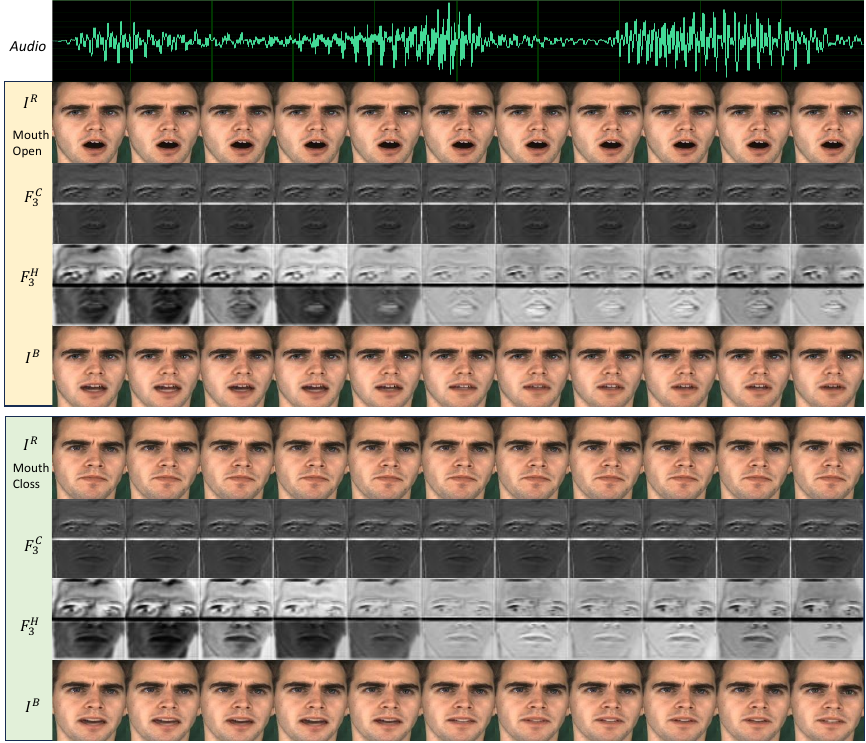}
	\vspace{-0pt}
	\caption{Effect of Reference Video on Results.	}
	\vspace{0pt}
	\label{fig:EffectReferenceVideo}
\end{figure*}

\subsection{Effect of Reference Video on Results}
The reference video carries the identity information of the driven character, which directly affects the character influence of the generated video. The expressions, poses, movements, and other information of the characters in the generated video are also directly obtained from the reference video. Therefore, the videos generated by our method naturally have more natural expressions, postures, and movements of the characters.

However, in addition to the above effects, the mouth shape of the reference video also has a greater effects on the mouth shape of the generated video, as shown in Fig.~\ref{fig:EffectReferenceVideo}, which we study in detail.
We choose a fixed frame as the input for the whole reference video $I^{R}$. The mouth of the person in the reference video in the yellow part is open, while the mouth of the person in the reference video in the green part is closed. 
We can see that at the same moment, the mouth of the $I^{B}$ obtained by using the open mouth as the reference video input is larger than the mouth of the $I^{B}$ obtained by using the closed mouth as the reference input. This phenomenon can also be seen in the $I_{3}^{H}$ feature map in FaceDecoder.
This is because the mouth shape state of the input reference video remains the same state after FaceEncoder (see $I_{3}^{C}$).
Therefore, it is not difficult to conclude that if the mouth in the reference video moves (i.e., talking) with timing and is inconsistent with the driving audio, it may cause the generated lips to be less lip-synchronous or cause lip-smacking.
Based on the above conclusions, if we want to get a better driving performance, we should choose a video in which the mouth does not move as a reference video.
If it is expected that the mouth of the person in the generated video will be wider, the mouth of the selected reference video must be kept wider.

\subsection{Additional Visualization Comparison}
\begin{figure*}[ht]
	\vspace{-1pt}
	\centering
	\includegraphics[width=1.0\linewidth]{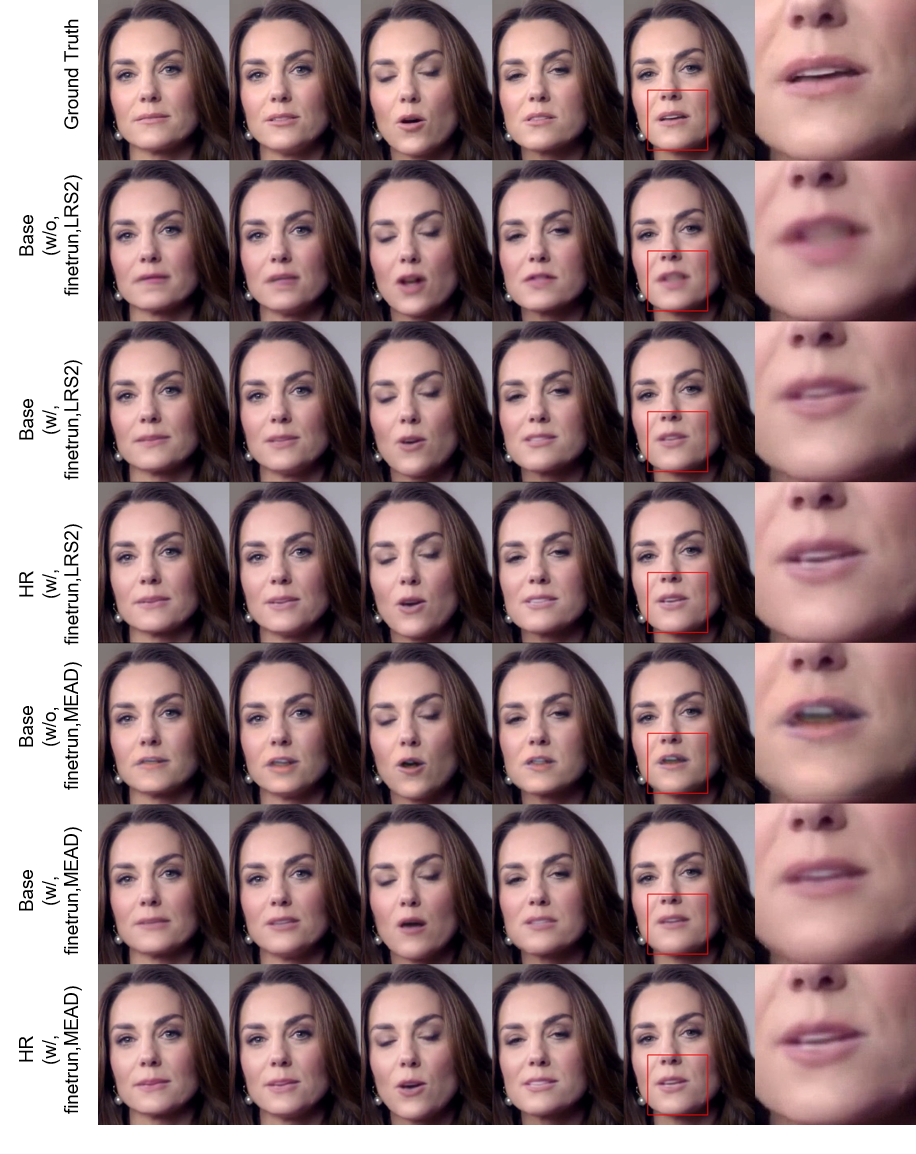}
	\vspace{-5pt}
	\caption{Visualization of Effectiveness for Finetuning.	}
	\vspace{0pt}
	\label{fig:katefineturn}
\end{figure*}

\textbf{Visualization of Effectiveness for Finetuning.} In our paper, we show the fine-tuned the pre-trained models on the MEAD and LRS2 datasets to Kate's videos and obtained corresponding results. Here, in Fig.~\ref{fig:katefineturn}, we show the Visualization of Effectiveness for Finetuning.
The results obtained by fine-tuning the pre-trained model on the data set of higher quality (e.g., MEAD-Neutral~\cite{wang2020mead}) are better and clearer than the results obtained by fine-tuning the pre-trained model on the data set of lower quality (e.g., LRS2~\cite{afouras2018deep}).

\begin{figure*}[ht]
	\vspace{-1pt}
	\centering
	\includegraphics[width=1.0\linewidth]{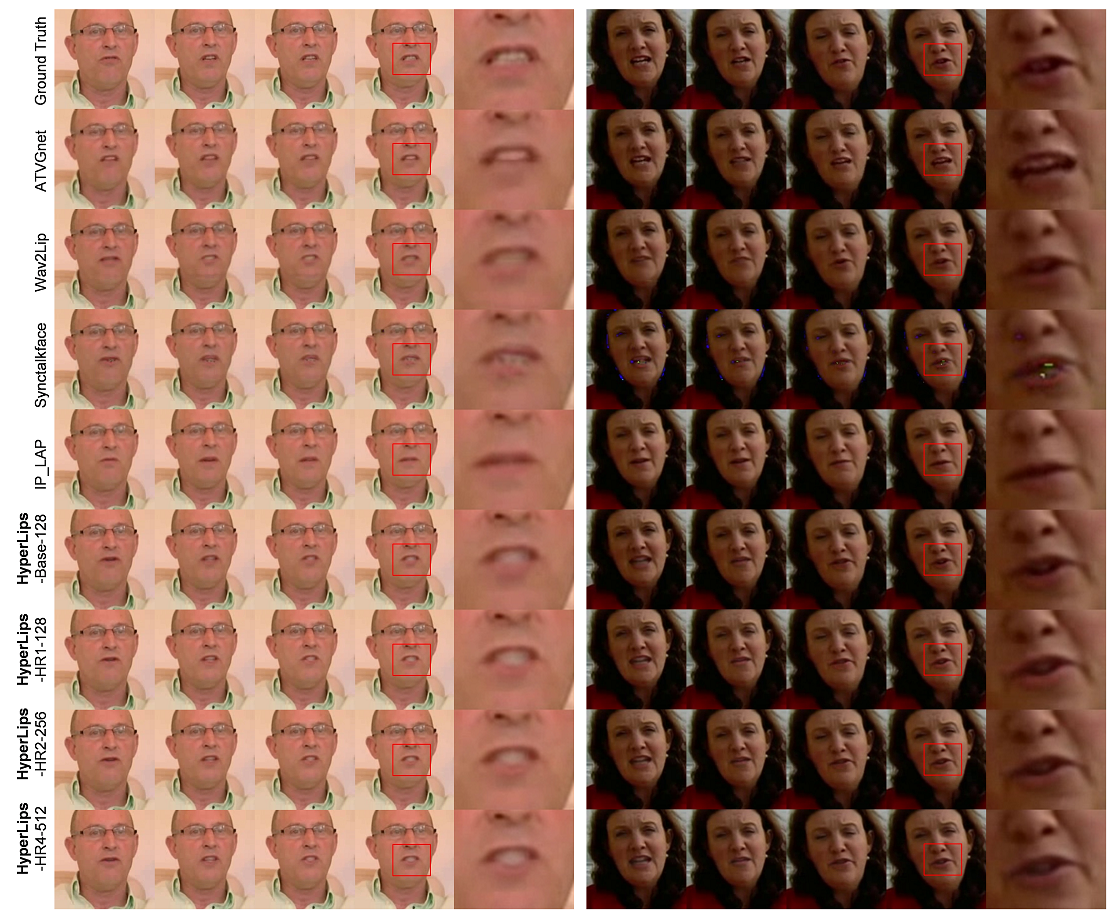}
	\vspace{-5pt}
	\caption{Qualitative comparisons with state-of-the-art methods on LRS2 \cite{afouras2018deep} datasets.	}
	\vspace{0pt}
	\label{fig:Qualitativelrs2s_lrs21}
\end{figure*}

\begin{figure*}[ht]
	\vspace{-1pt}
	\centering
	\includegraphics[width=1.0\linewidth]{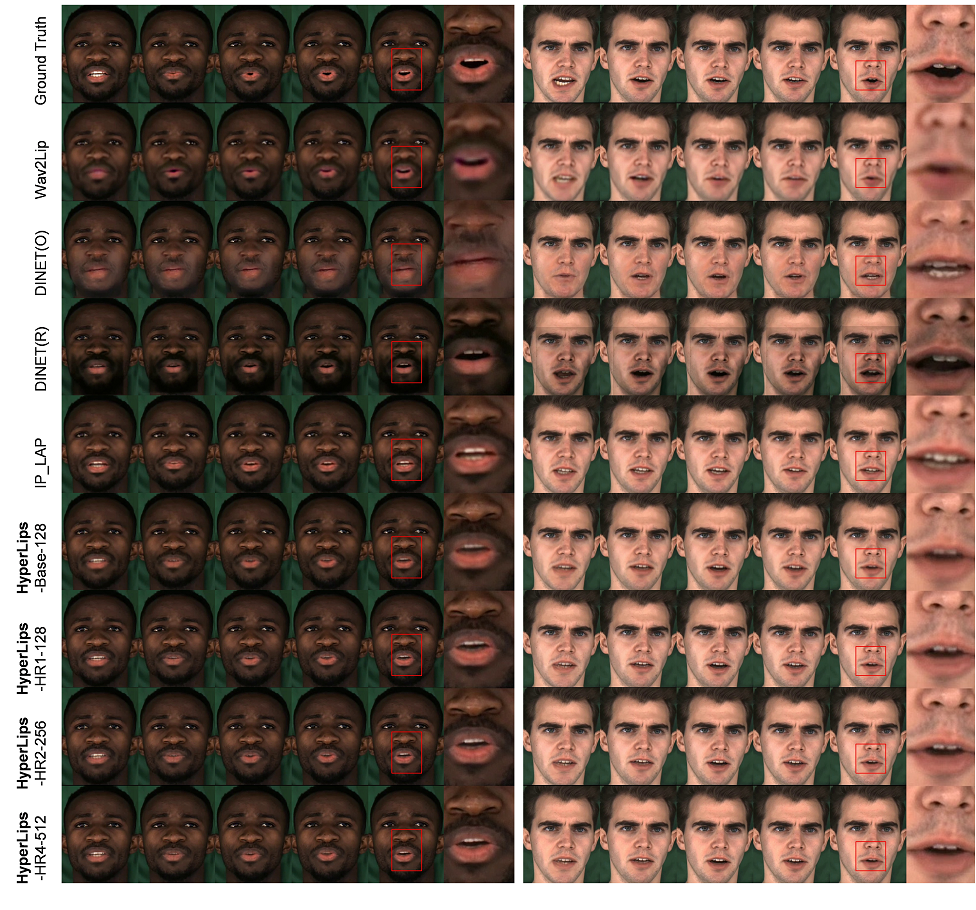}
	\vspace{-5pt}
	\caption{Qualitative comparisons with state-of-the-art methods on MEAD-Neutral \cite{wang2020mead} datasets.	}
	\vspace{0pt}
	\label{fig:Qualitativelrs2s_s_mead1}
\end{figure*}

\begin{figure*}[ht]
	\vspace{-1pt}
	\centering
	\includegraphics[width=1.0\linewidth]{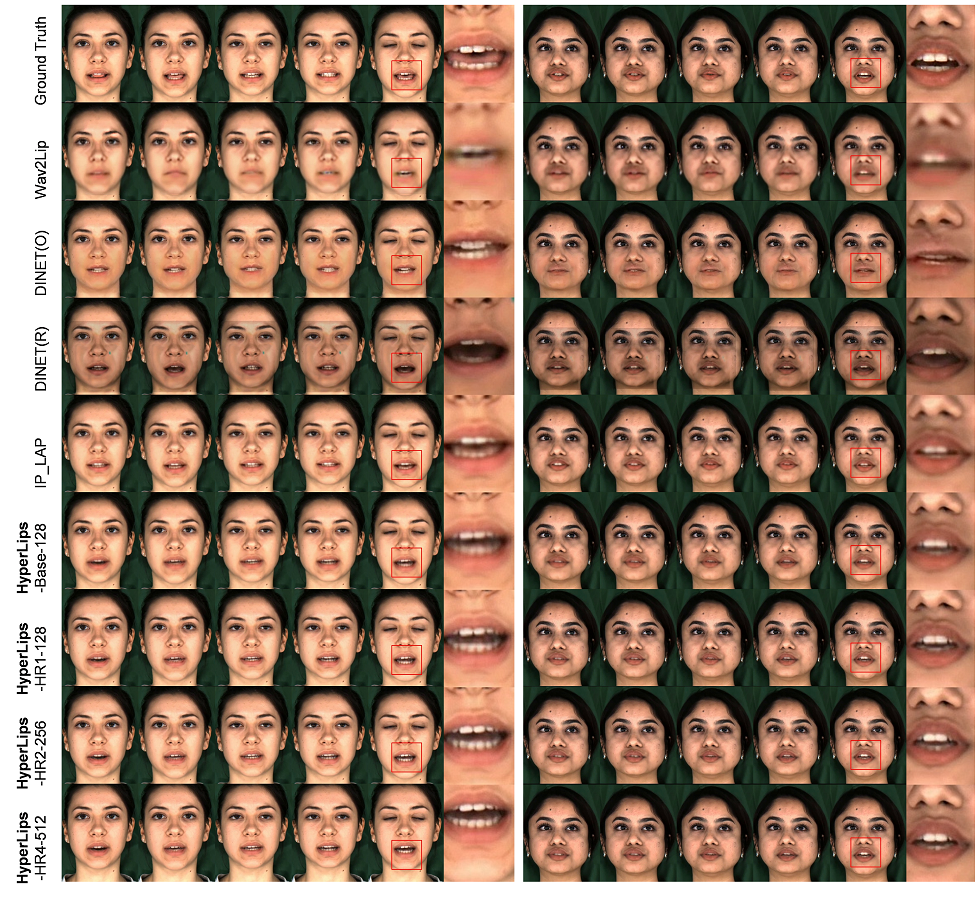}
	\vspace{-5pt}
	\caption{Qualitative comparisons with state-of-the-art methods on MEAD-Neutral \cite{wang2020mead} datasets.	}
	\vspace{0pt}
	\label{fig:Qualitativelrs2s_s_mead2}
\end{figure*}

\begin{figure*}[ht]
	\vspace{-1pt}
	\centering
	\includegraphics[width=1.0\linewidth]{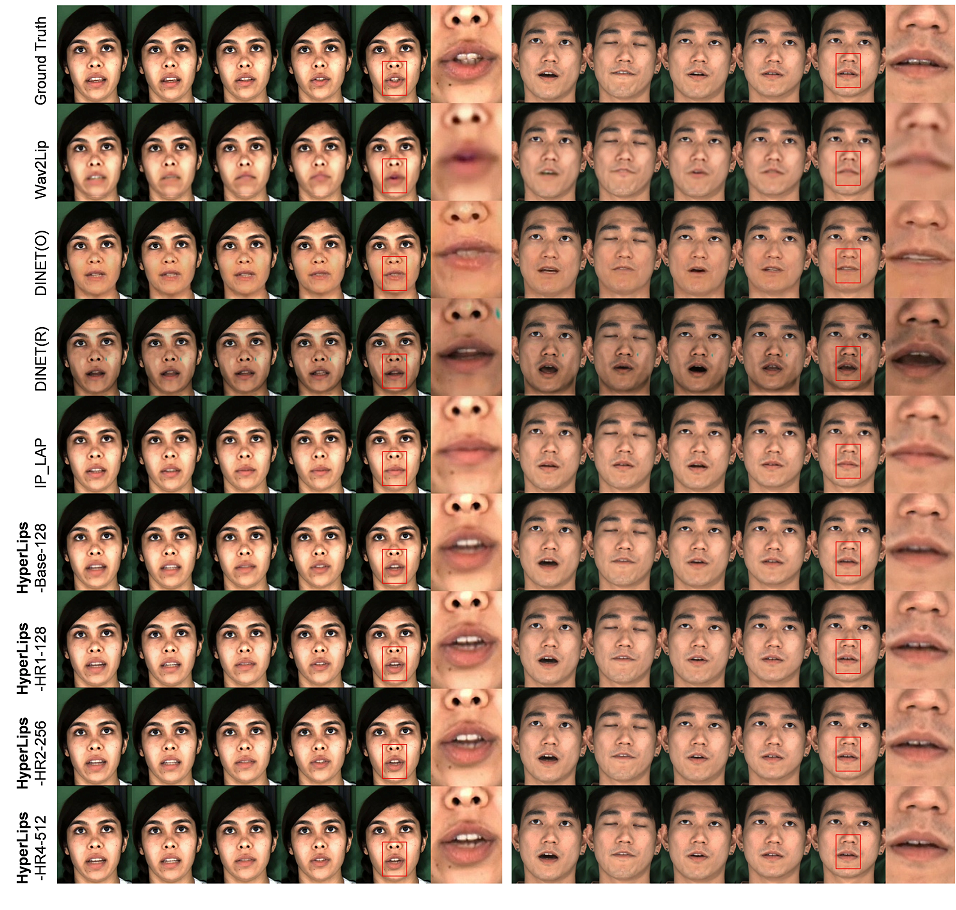}
	\vspace{-5pt}
	\caption{Qualitative comparisons with state-of-the-art methods on MEAD-Neutral \cite{wang2020mead} datasets.	}
	\vspace{0pt}
	\label{fig:Qualitativelrs2s_s_mead3}
\end{figure*}

\textbf{Visualization Comparison with State-of-the-art}: We show more visualization comparisons on the LRS2 and MEAD datasets in Fig.~\ref{fig:Qualitativelrs2s_lrs21},  Fig.~\ref{fig:Qualitativelrs2s_s_mead1},  Fig.~\ref{fig:Qualitativelrs2s_s_mead2},  Fig.~\ref{fig:Qualitativelrs2s_s_mead3}, etc. These comparisons include our base model as well as all HR models. The results show that our method outperforms in terms of visual quality and lip synchronization. This excellence is more evident on high-definition datasets (such as the MEAD dataset).

\vfill

\end{document}